\documentclass[preprint,12pt]{elsarticle}

\usepackage{amssymb}
\usepackage{amsmath}
\usepackage{amsthm}
\usepackage{caption}
\usepackage{derivative}
\usepackage{multirow}
\usepackage{booktabs}
\usepackage{comment}
\usepackage{lineno}
\usepackage{adjustbox}
\usepackage{standalone}
\usepackage{float}
\usepackage{url}
\usepackage[titletoc]{appendix}

\usepackage{xcolor}
\usepackage{color}
\newcommand{\rr}{\color{black}}

\usepackage{setspace}

\journal{Computerized Medical Imaging and Graphics}

\begin{document}

\begin{frontmatter}

\title{XGeM: A Multi-Prompt Foundation Model for Multimodal Medical Data Generation}

\author[UCBM]{Daniele Molino} 
\ead{daniele.molino@unicampus.it}

\author[UMU]{Francesco Di Feola} 
\ead{francesco.feola@umu.se}

\author[policlinico1,policlinico2]{Eliodoro Faiella}
\ead{e.faiella@policlinicocampus.it}

\author[CDI]{Deborah Fazzini}
\ead{deborah.fazzini@CDI.it}

\author[policlinico1]{Domiziana Santucci}
\ead{d.santucci@policlinicocampus.it}

\author[Shen]{Linlin Shen}
\ead{llshen@szu.edu.cn}

\author[UCBM]{Valerio Guarrasi\fnref{equal}} 
\ead{valerio.guarrasi@unicampus.it}

\author[UCBM,UMU]{Paolo Soda\corref{cor1}\fnref{equal}} 
\ead{p.soda@unicampus.it, paolo.soda@umu.se}
\cortext[cor1]{Corresponding author: p.soda@unicampus.it, paolo.soda@umu.se}

\fntext[equal]{These authors equally contributed to the work and share senior authorship.}

\affiliation[UCBM]{organization={Unit of Artificial Intelligence and Computer Systems, Department of Engineering, Università Campus Bio-Medico di Roma},
            city={Roma},
            country={Europe}}

\affiliation[UMU]{organization={Department of Diagnostics and Intervention, Biomedical Engineering and Radiation Physics, Umeå University},
            city={Umeå},
            country={Sweden}}

\affiliation[CDI]{organization={Department of Diagnostic Imaging and Stereotactic Radiosurgey, Centro Diagnostico Italiano S.p.A.},
                  city={Milano},
                  country={Italy}}

\affiliation[policlinico1]{organization={Department of Radiology and Interventional Radiology, Fondazione Policlinico Universitario Campus Bio-Medico},
city={Rome},
country={Italy}}

\affiliation[policlinico2]{organization={Research Unit of Radiology and Interventional Radiology, Department of Medicine and Surgery, Università Campus Bio-Medico di Roma},
city={Rome},
country={Italy}}

\affiliation[Shen]{organization={College of Computer Science and Software Engineering, Shenzhen University},
                    city={Shenzhen},
                    country={China}}

\begin{abstract}
The adoption of Artificial Intelligence in medical imaging holds great promise, yet it remains hindered by challenges such as data scarcity, privacy concerns, and the need for robust multimodal integration.
While recent advances in generative modeling have enabled high-quality synthetic data generation, existing approaches are often limited to unimodal, unidirectional synthesis and therefore lack the ability to jointly synthesize multiple modalities while preserving clinical consistency.
To address this challenge, we introduce XGeM, a 6.77-billion-parameter multimodal generative model designed to support flexible, any-to-any synthesis between medical data modalities.
XGeM constructs a shared latent space via contrastive learning and introduces a novel Multi-Prompt Training strategy, enabling conditioning on arbitrary subsets of input modalities. 
This design allows the model to adapt to heterogeneous clinical inputs and generate multiple outputs jointly, preserving both semantic and structural coherence.
We extensively validate XGeM: first we benchmark it against five competitors on the MIMIC-CXR dataset, a state-of-the-art dataset for multi-view Chest X-ray and radiological report generation.
Secondly, we perform a Visual Turing Test with expert radiologists to assess the realism and clinical relevance of the generated data, ensuring alignment with real-world scenarios.
Finally, we show how XGeM can support key medical data challenges such as anonymization, class imbalance, and data scarcity, underscoring its utility as a foundation model for medical data synthesis.
Project page is at \url{https://cosbidev.github.io/XGeM/}.
\end{abstract}

\end{frontmatter}

\section{Introduction}
\label{sec1}
\noindent
Artificial Intelligence (AI) is increasingly transforming healthcare by enabling automated, data-driven approaches to enhance diagnostics, prognosis, and treatment planning.
Today, AI systems can process large-scale medical data to uncover subtle patterns often imperceptible to clinicians, supporting diagnostic precision and personalized care~\cite{alowais2023revolutionizing}.
The integration of multimodal data offers a major step forward, enabling richer diagnostic insights through the fusion of heterogeneous medical information~\cite{guarrasi2025systematic}.
However, despite these advancements, the implementation of AI in healthcare faces several challenges, primarily caused by data scarcity and privacy concerns~\cite{alzubaidi2023survey}.
Although essential for safeguarding patient information, privacy regulations such as the GDPR~\cite{gdpr2016general} in Europe and HIPAA~\cite{act1996health} in the United States often hinder collaborative data sharing across institutions.
As a consequence, the available datasets for training AI models are limited in size, diversity, and scope, posing a substantial challenge for deep learning (DL) models, which typically require large, high-quality datasets to generalize effectively.
Furthermore, with the absence of sufficient diversity, models tend to overfit or exhibit bias, hindering their clinical reliability.
These challenges have led to growing interest in synthetic data generation, with multimodal approaches offering a way to capture richer clinical variability.
This emerging approach involves creating artificial data that replicate the complexity and diversity of real medical data, thus providing a solution to bypass the constraints of real-world scarcity and privacy concerns.

\subsection{Generative AI}
\noindent
Generative AI has advanced significantly since the introduction of Generative Adversarial Networks (GANs)~\cite{goodfellow2020generative} in 2014, which enabled the generation of synthetic data through adversarial training.
However, GANs suffer from limitations such as training instability, mode collapse, and difficulty in modeling fine-grained details, factors that hinder their applicability in medical imaging~\cite{saad2024survey}.
More recently, Diffusion Models (DMs)~\cite{ho2020denoising} have emerged as a powerful alternative for data generation.
By iteratively denoising noise-corrupted representations, diffusion models can generate high-fidelity and diverse samples, effectively capturing the subtle anatomical details required in medical imaging applications.
Latent Diffusion Models (LDMs)~\cite{rombach2022high} perform the denoising process in a compressed latent space, significantly reducing computational requirements and enabling broader applicability across resource-constrained settings.
Furthermore, their flexible conditioning mechanisms enable fine-grained control over the generation process~\cite{ramesh2022hierarchical, saharia2022photorealistic}.
This is achieved through modality-specific encoders that extract semantic representations from the input, guiding the synthesis of anatomically or clinically relevant features.
Such controllability enhances both model flexibility and clinical applicability, allowing synthetic data to be tailored to specific diagnostic or research contexts.
With appropriate pretraining, LDMs can be effectively adapted for a variety of downstream generation tasks, highlighting their strong potential as foundation models for medical data synthesis~\cite{bommasani2021opportunities}.
However, most existing models are limited to single-modality translation, which constrains their utility in clinical workflows which inherently rely on the integration of multiple data modalities.
Outside the medical domain, significant advancements have been made in multimodal data generation. 
A notable example is CoDi~\cite{tang2023anytoany}, which supports any-to-any generation by leveraging a shared latent space constructed via self-supervised contrastive learning across input modalities. 
This approach enhances cross-modal consistency and coherence, allowing for multimodal generation while avoiding the pitfalls of multi-stage approaches, which often suffer from error propagation~\cite{li2023error}.
The adaptation of a similar approach for medical data generation could prove highly beneficial, filling a critical gap in the availability of diverse and high-quality datasets for research and diagnostic purposes.
However, CoDi presents some limitations when applied to such a setting: while it demonstrates the feasibility of any-to-any generation in non-medical environment, its performance tends to degrade when provided with multiple input modalities or their combinations, a limitation that cannot be overlooked in the medical domain, where reliability and consistency across modalities are critical.

\subsection{Related Works}
\label{sec2}
\noindent
In recent years, Chest X-rays (CXRs) have been the focus of generative modeling efforts in the medical domain, due to their potential to provide insights into a wide range of medical conditions.
Among these, the task of generating radiology reports from X-ray images has received the most attention, leading to the development of methods capable of producing fluent and clinically informative narratives from visual inputs~\cite{chen2020generating, nicolson2024longitudinal, miura2020improving}.

In parallel, a smaller but growing line of research has explored the reverse task, i.e., generating radiographic images from textual descriptions.
Early works primarily relied on GANs, often constrained to narrow clinical contexts such as tuberculosis~\cite{moris2022unsupervised}, pneumonia~\cite{srivastav2021improved}, or COVID-19~\cite{karbhari2021generation, shams2020generative}. 
More recently, LDMs have emerged as a promising alternative for CXR generation.
RoentGen~\cite{chambon2022roentgen} was the first work to adapt a pretrained LDM for text-conditioned CXR generation beyond few- or zero-shot setting~\cite{chambon2022adapting, packhauser2023generation}. 
To effectively transfer the model to the medical domain, they demonstrated that fine-tuning the UNet component is essential, as it enables to capture the unique features and nuances of medical images.

While these advancements have led to notable improvement in both report generation from images and image synthesis from text, addressing these tasks independently constrains the ability of generative models to capture cross-modal dependencies, essential for clinically coherent generation.
In clinical practice, visual and textual information are inherently complementary; therefore, the independent generation of a single modality often results in semantic inconsistencies and fails to fully exploit the richness of multimodal representations.

To address this limitation, recent efforts have proposed models capable of bidirectional generation, synthesizing both radiographic images and textual reports within a unified framework.
In this context, UniXGen~\cite{lee2023unified} introduced a transformer-based~\cite{vaswani2017attention, choromanski2020rethinking} approach for both CXR and report generation. 
They utilize VQ-GAN~\cite{esser2021taming} to discretize X-rays into token sequences, allowing both image and text generation to be formulated as sequence modeling tasks. 
Notably, UniXGen supports multi-view synthesis, leveraging the complementary information provided by the frontal and lateral X-ray projections.
However, its architecture requires separate processing for each modality and lacks explicit mechanisms to ensure inter-modality output consistency.

Lee et al.~\cite{lee2023llm} proposed LLM-CXR, a large language model fine-tuned for the bidirectional generation of CXR and radiology reports. 
Similar to UniXGen, this approach tokenizes X-ray images into discrete representations, enabling the model to perform both image and text generation within a unified autoregressive framework.
However, LLM-CXR is limited to single-output generation, requiring separate inferences for each modality, and operates exclusively on frontal chest views, thereby reducing its applicability in more comprehensive diagnostic scenarios.

Despite their contributions, these approaches overlook the complementary nature of different medical data modalities and are unable to generate multimodal outputs simultaneously.
This independent processing often leads to inconsistencies between modalities, resulting in outputs that may lack clinical coherence.
Such limitations hinder their applicability in real-world healthcare scenarios, where the seamless integration of multimodal information is crucial to accurately reflect the complexity of patient data.

\subsection{Contribution}
\noindent
We introduce XGeM, a multimodal generative architecture for medical data synthesis that builds upon the CoDi framework through domain-specific adaptation.
Leveraging contrastive learning, XGeM constructs a shared latent space that supports flexible, any-to-any generation between heterogeneous modalities, such as frontal/lateral X-rays and radiology reports.
To improve cross-modal integration, we propose a novel Multi-Prompt Training strategy, which dynamically fuses multiple conditioning signals at training time, enabling the model to adapt to realistic clinical scenarios where available data may vary across patients and modalities.
\\
The main contribution can be summarized as:
\begin{itemize}
    \item We present XGeM, a modular foundation model capable of synthesizing multiple medical data modalities from a shared latent space, supporting flexible any-to-any generation.
    \item We introduce a novel Multi-Prompt Training mechanism that, by dynamically combining available modality embeddings into a unified conditioning vector during training, allows the model to synthesize on arbitrary combinations of modalities, thus improving generalization across diverse input configurations.
    \item We thoroughly evaluate XGeM against existing state-of-the-art models, showing its superior capabilities in terms of quality, realism and clinical accuracy.
    \item We perform a Visual Turing Test, consisting of five evaluation tasks administrated to three expert radiologists, to assess the clinical realism and diagnostic consistency of the generated data modalities.
    \item We evaluate the utility of synthetic data, by showing its effectiveness in tackling three key challenges in the medical domain, i.e., Anonymization, Imbalance Learning and Data Scarcity.
\end{itemize}
This paper is organized as follows: Section 2 presents the methods employed in this work, detailing the architecture of XGeM, its training procedure, and the innovative Multi-Prompt Training strategy. 
Section 3 describes the dataset and preprocessing steps.
It then outlines the experimental setup, including the competitors, evaluation metrics, and configurations adopted to assess the performance of XGeM. 
Section 5 discusses the results, providing both quantitative and qualitative analyses, also introducing the findings from the Visual Turing Test. 
Finally, Section 6 summarizes the main contributions and discusses potential directions for future research.

\section{Methods}
\label{sec3}
\noindent
Assuming ${\mathcal{M}}$ is the set of our modalities, let ${\mathcal{I}}=\{I_{1}, I_{2}, ..., I_{n}\}$ be any subset of modalities used as prompts for the generation and let ${\mathcal{O}}=\{O_{1}, O_{2}, ..., O_{m}\}$ be any subset of modalities to be generated, such that $\mathcal{O} \cap \mathcal{I} = \varnothing$, with ${\mathcal{I}}, {\mathcal{O}} \subseteq {\mathcal{M}}$.
This separation is introduced for clarity of exposition; in practice, any modality can serve as either input or output, and XGeM supports arbitrary combinations thereof.
\begin{figure*}[h]
\centering
\includegraphics[width=1\textwidth]{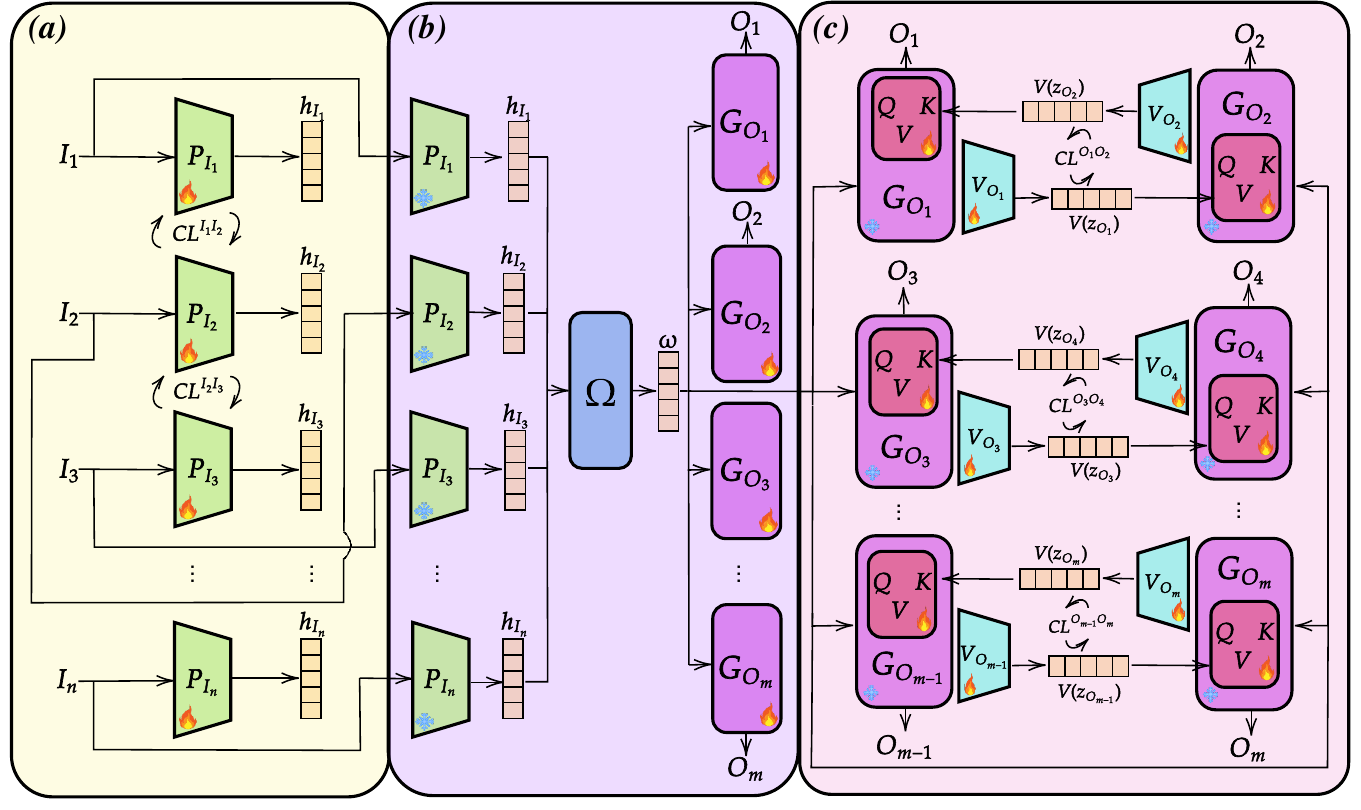}
\captionsetup{justification=raggedright, singlelinecheck=false}
\caption[Framework of XGeM]{\textbf{Framework of XGeM} - a) Shared Latent Space construction: Input modalities are processed by modality-specific prompt encoders to extract feature representations, which are aligned using contrastive learning. - b) Single modality generation training: Individual LDMs are trained for each output modality using the proposed Multi-prompt training approach. This technique dynamically combines subsets of input modalities to form a conditioning vector, allowing the model to learn from various input configurations. - c) Latent Cross-Modal Alignment: This phase enable simultaneous multimodal generation, enabling mutual conditioning between LDMs.}\label{fig:model}
\end{figure*}
\\
The XGeM framework, illustrated in Fig.~\ref{fig:model}, comprises three main components, each corresponding to a distinct phase in the training pipeline.
Panel (a) shows how modality-specific prompt encoders map input data into a shared latent space via contrastive learning.
Panel (b) illustrates the training of individual LDMs for each output modality, conditioned through the proposed Multi-Prompt strategy.
Panel (c) introduces a cross-modal alignment mechanism, which enables simultaneous generation of multiple modalities by allowing mutual information flow between LDMs.
The following sections detail each component of the training pipeline: latent space construction (Section~\ref{Prompt Encoder Training}), single-modality generation via Multi-Prompt Training (Section~\ref{Multi-Prompt Training}), and multi-output generation through latent cross-modal alignment (Section~\ref{Multi-Output Generation}).

\subsection{Building a Shared Latent Space}
\label{Prompt Encoder Training}
\noindent
To enable flexible conditioning across heterogeneous inputs, we align all modality-specific representations within a shared latent space using contrastive learning.
This alignment enables the extraction of coherent representations from any input modality, providing the foundation for conditioning the generation process on diverse input configurations.

Following the Bridging Alignment strategy introduced in~\cite{tang2023anytoany}, we enforce consistency among modality-specific representations through pairwise training.
As shown in Fig.~\ref{fig:model}.a, each input modality $I_j \in \mathcal{I}$ is encoded into a feature vector $h_{I_j}=P_{I_j}(I_j)$ via its modality-specific prompt encoder $P_{I_j}$.
Given two modalities $I_j$, $I_{j+1}$, their corresponding embeddings $h_j$, $h_{j+1}$ are trained to be close if derived from the same sample and distant otherwise.
To this end, we leveraged the InfoNCE contrastive loss~\cite{oord2018representation}:
\begin{equation}
    {{\cal L}_{I_j,I_{j+1}} =  - \log \frac{{\exp ({h_{I_j}^a}^{\top} \cdot {h_{I_{j+1}}^a}/\tau )}}{{\exp ({h_{I_j}^a}^{\top}\cdot{h_{I_{j+1}}^a}/\tau ) + \sum\nolimits_{b \ne a} {\exp ({h_{I_j}^b}^{\top}\cdot{h_{I_{j+1}}^b}/\tau )}}}}
    \label{eq:infonce}
\end{equation}
where $\tau$ is the scalar temperature regulating the softness of the softmax distribution, and $a$, $b$ refers, respectively, to positive and negative couples. 
We adopt the symmetric loss ${\mathcal{L}_{I_j, I_{j+1}}}+{\mathcal{L}_{I_{j+1}, I_j}}$ to make the embeddings closer together.

The shared latent space is built through iterative pairwise contrastive training, which ensures multimodal alignment while maintaining computational efficiency.

\subsection{Multi-prompt Training for single-modality generation}
\label{Multi-Prompt Training}
\noindent
Training a generative model capable of handling multiple input and output modalities demands exposure to diverse modality combinations while maintaining consistent output quality.
To address this, XGeM adopts a modular architecture, where each modality-specific LDM is trained independently and later integrated into a unified generation framework.
Since medical data is inherently multimodal and often available in partial combinations~\cite{caruso2024maria}, we introduce a novel training approach, named Multi-Prompt Training, which simulates such variability by dynamically varying the conditioning inputs at training-time.
This strategy improves the model's ability to incorporate information from multiple input modalities and remain robust to varying input configurations.

Recall that $O_i$ denotes the target modality to be generated, and $\mathcal{I}$ the set of conditioning inputs.
As shown in Fig.\ref{fig:model}.b, we extract the latent representations $h_{I_j} = P_{I_j}(I_j)$ for each $I_j \in \mathcal{I}$ using the pretrained and now frozen prompt encoders from Section~\ref{Prompt Encoder Training}.
Then, at each training iteration, a sampling strategy $\Omega$ selects a random non-empty subset $I_p \subset \mathcal{I}$, whose representations are combined to form the conditioning vector $\omega$.
Given $n$ available modalities, there exists $2^{n-1}-1$ possible combinations, making the probability of drawing any possible subset equal to $p = \frac{1}{2^{(n-1)} - 1}$. 
This uniform sampling ensures equal exposure to all possible conditioning subsets, encouraging the model to generalize across diverse multimodal contexts.

Once a combination is selected, their latent representations are linearly interpolated to form a conditioning vector defined as:
\begin{equation}
\label{eq:prompt_sample}
\omega = \Omega(h_{I_1}...h_{I_n}) = \sum_{j=1}^n \alpha_j h_{I_j} \text{ with } \sum_{j=1}^n \alpha_j = 1 \text{ and } j \in \{I_p\}.
\end{equation}
\\
The resulting vector $\omega$ is then used as the conditioning input for $G_{O_i}$.

Following the denoising diffusion formulation from~\cite{ho2020denoising, rombach2022high}, the training loss for $G_{O_i}$ is defined as:
\begin{equation}
\label{eq:df}
    \mathcal{L}_{D}^{O_i} = \mathbb{E}_{z, \epsilon, t} \left[ \| \epsilon - \epsilon_{\theta}(z_{O_i}, t, \omega) \|_2^2 \right]
\end{equation}
Here, $z_{O_i}$ denotes the latent representation at diffusion timestep $t \in [1, T]$, sampled uniformly, and $\epsilon_\theta$ is the denoising UNet parameterized by $\theta$.

This training strategy is independently applied to every output modality, allowing each generator to be flexibly conditioned on varying subsets of input data.

\subsection{Multi-output generation via Cross-modal Latent Alignement}
\label{Multi-Output Generation}
\noindent
The final training stage enables joint generation of multiple output modalities by introducing mutual conditioning mechanisms between the previously trained modality-specific generators.

To this end, we incorporate two trainable components into each LDM $G_{O_i}$:
the first is a projection encoder ${V_{O_i}}$, that maps the latent variable $z_{O_i}$ into a shared latent space; the second is a cross-attention layer, which allows each LDM to attend to the latent states of other generation processes.

Let $O_i$ and $O_{i+1}$ be two modalities jointly generated by $G_{O_i}$ and $G_{O_{i+1}}$, with corresponding latent variables $z_{O_i}$ and $z_{O_{i+1}}$ at a given diffusion step.
As illustrated in Fig.~\ref{fig:model}.c, $V_{O_{i+1}}$ projects $z_{O_{i+1}}$ into a shared latent space, and this representation is injected via cross-attention into $G_{O_i}$.
\\
For the diffusion model of modality $O_i$, the training objective in Eq.\ref{eq:df} become: 
\begin{equation}
\mathcal{L}_{D}^{O_i} = \mathbb{E}_{z, \epsilon, t}\|\epsilon - \epsilon_{\theta_c}(z_{O_i}, V_{O_{i+1}}(z_{O_{i+1}}), t, \omega)\|_2^2,
\end{equation}
Here, $\theta_c$ denotes the parameters of the cross-attention-augmented UNet.
\\
The training objective for the joint generation of $O_i$ and $O_{i+1}$ becomes $\mathcal{L}_{D}^{O_i}+\mathcal{L}_{D}^{O_{i+1}}$.
\\
To further enhance the training process, we incorporated contrastive learning between the outputs of the projection encoders, using again the InfoNCE contrastive loss, as defined in Eq.\ref{eq:infonce}.

By training only the projection encoder and cross-attention modules while freezing the LDM backbones, XGeM effectively learns to jointly generate modalities while maintaining high-quality outputs.

\section{Experimental Configuration}
\label{sec5}
\noindent
This section provides a comprehensive overview of the experimental setup adopted to evaluate the performance of XGeM. 
It begins by describing the dataset used, including its key characteristics and preprocessing steps. 
Next, the section details the implementation specifics of the proposed framework, such as architectural choices, training configurations, and optimization strategies. 
Additionally, it presents the competitors used for comparative analysis, highlighting their relevance and limitations in the context of multimodal medical data generation. 
Finally, the evaluation metrics are introduced, encompassing both quantitative and qualitative measures to thoroughly assess the realism, coherence, and clinical utility of the generated outputs.

\subsection{Materials}
\label{sec4}
\noindent
We used the MIMIC-CXR~\cite{johnson2019mimic} dataset that contains 377,110 CXR images along with their corresponding radiology reports, for a total of 227,827 studies conducted at the Beth Israel Deaconess Medical Center in Boston, MA, USA.
Each study includes X-ray images acquired in both frontal and lateral projections.
These two views provide distinct but complementary anatomical perspectives, enhancing diagnostic comprehensiveness~\cite{santosh2018angular}.
The lateral view mitigates occlusions caused by cardiovascular structures in the frontal projection, responsible for hiding up to 15\% of the lung volume, thus revealing lesions otherwise undetectable~\cite{raoof2012interpretation, hashir2020quantifying}.
Therefore, we model frontal and lateral CXRs as distinct modalities within our framework.
Each radiology report consists of two sections: a finding section that provides a detailed description of both normal and abnormal features observed in the corresponding CXR, and an impression section, which provides a concise summary of the findings intended to support medical decision-making.
In this work, we focused exclusively on the latter, as it offers a concise yet powerful summary of the patient’s condition.
\\
From the repository, we extracted a total of 154,721 X-rays from 78,584 studies, retaining only cases that included a radiology report and both frontal and lateral views.
We relied on the original DICOM files~\cite{bidgood1997understanding}, as subtle image details are essential in medical imaging and could be lost through compression.
\\
X-ray preprocessing included multiple steps to standardize image format and resolution for model training.
We first examined the pixel spacing of each image and resampled those with non-standard values to $[0.139, 0.139]$, the most frequent spacing found in 95.76\% of the dataset.
Images with a Photometric Interpretation of Monochrome-1 were inverted to ensure consistent intensity representation.
We normalized the images to the $[0, 1]$ range by dividing each pixel by the maximum value allowed by the image’s bit depth.
To preserve anatomical proportions, we avoided direct resizing of the non-square scans, which could distort spatial structures.
Cropping was also ruled out, as it was not feasible to define a consistent Region of Interest (ROI) applicable across all scans.
Instead, we applied zero-padding and resized the images to $256\times256$, standardizing input size while preserving anatomical integrity.
\\
To ensure an unbiased evaluation, we extracted a holdout test set before any training procedure. 
This set consists of 33.588 samples, that were selected to guarantee no overlap between training and test samples.

\subsection{Model Architecture and Training}
\label{Model Training}
\noindent
We delve here into the architectural choices made for each component of our framework, outlining their design rationale and integration within the overall system. 

\subsubsection{Prompt Encoders}
\noindent
Given that our three modalities consist of texts and images, we adopt the Contrastive Language-Image Pretraining (CLIP)~\cite{radford2021learning} approach to leverage a pretrained text-image paired encoder. 
This approach is composed of an image and a text encoder, denoted as $P_X$ and $P_R$, jointly trained on large-scale datasets of text-image pairs. 
In order to reduce the computational overload, we decided to let a single image encoder, i.e. a ViT Transformer, be responsible for both frontal and lateral X-rays feature representations, while we leverage a masked self-attention Transformer for the text encoding.
Training was performed using the Adam optimizer with a learning rate of $5 \times 10^{-5}$, weight decay of  $1 \times 10^{-4}$ and batch size of 1024 for 200 epochs.

\subsubsection{Latent Diffusion Model}
\noindent
\textbf{X-ray Diffusion Model:}
Our image generation module employs a frozen AutoKL variational autoencoder~\cite{esser2021taming} to map high-resolution chest X-rays into a compact latent space~\cite{rombach2022high}.
Following~\cite{chambon2022adapting}, we fine-tune only the UNet component of the diffusion model, which has proven effective for adapting LDMs to medical imaging tasks.
Two separate models were trained for frontal and lateral X-rays, using a batch size of 512, a learning rate of $1 \times 10^{-5}$, and a weight decay of $1 \times 10^{-4}$.
Both models were trained for 100 epochs using the AdamW optimizer.
\noindent
\\
\textbf{Report Diffusion Model:}
For text generation, we leveraged Optimus~\cite{li2020optimus} as the text VAE, comprising a BERT encoder~\cite{devlin2018bert} and a GPT-2 decoder~\cite{radford2019language}
The UNet design follows~\cite{xu2023versatile}, leveraging fully connected residual blocks, which expand the 768-dimensional latent vectors into a 320×4 feature map, using GroupNorm~\cite{wu2018group}, SiLU~\cite{elfwing2018sigmoid}, and skip connections.

In contrast to X-ray generation, where only the UNet is adapted, we fine-tune both the VAE and UNet for text generation. 
This is necessary to capture the specialized structure and vocabulary of radiology reports, which differ significantly from the textual data seen during pretraining.
Following~\cite{li2020optimus}, the training process begins with a reconstruction task, where the VAE is tasked with accurately reconstructing input radiology reports from their latent representations.
We train the VAE for 100 epochs using a batch size of 256 and the AdamW optimizer with a learning rate of $5 \times 10^{-5}$ and weight decay of $1 \times 10^{-4}$

Once the first step is fulfilled, the UNet is trained for report generation with a batch size of 1024 and a learning rate of $1 \times 10^{-5}$. 
The weight decay, the optimizer configuration and the number of epochs remained consistent with those used for the X-ray LDMs.

\subsection{Computational Analysis} 
\noindent 
To quantify the computational cost of our framework, we provide a detailed breakdown of the number of parameters for each model component. 
Specifically, CLIP model contains 737 million parameters, while AutoKL has 101 million parameters, with two instances used in our framework. 
Optimus model consists of 241 million parameters, and the X-ray UNet model has 1.77 billion parameters, with two instances used. 
Finally, the Report UNet model has 2.04 billion parameters. 
In total, the number of parameters for all components combined amounts to 6.77 billion.
All experiments were conducted on a high-performance computing cluster equipped with four NVIDIA A100 GPUs. 
The total computational time required across all experiments was approximately 38,354 hours. 

\subsection{Competitors}
\noindent
To enable a rigorous comparison, we selected six representative baselines for X-ray and report generation, each with publicly available code and weights and pretrained on MIMIC-CXR, thus ensuring reproducibility and enabling direct comparison without further fine-tuning.
UniXGen and LLM-CXR~\cite{lee2023unified, lee2023llm} were included as they tackle bidirectional X-ray and report generation using architectures distinct from ours, specifically Transformers and Large Language Models.
In contrast, although RoentGen~\cite{chambon2022roentgen} and CXRMate~\cite{nicolson2024longitudinal} do not support bidirectional generation, they were included to provide a comparison with task-specific models that focus exclusively on either X-ray or report generation.

\begin{itemize}
    \item \textbf{CoDi}~\cite{tang2023anytoany}: a general-purpose multimodal diffusion model used without any adaptation to the medical domain.
    \item \textbf{$\text{CoDi}_{\mathit{{XR}}}$}~\cite{molino2025any}: an ablated version of XGeM without the Multi-Prompt Training.
    This variant highlights the importance of the approach in enhancing the model's capabilities to integrate information from multiple input modalities.
    \item \textbf{UniXGen}~\cite{lee2023unified}: a Transformer-based model that formulates both X-ray and report generation as sequence prediction tasks via visual tokenization.
    It supports multi-view generation but handles each output modality independently.
    \item \textbf{LLM-CXR}~\cite{lee2023llm}: a fine-tuned large language model for CXR–report generation.
    It supports bidirectional generation but is limited to frontal images.
    \item \textbf{RoentGen}~\cite{chambon2022roentgen}, a text-conditioned Stable Diffusion~\cite{rombach2022high} fine-tuned for generating synthetic frontal CXRs based on textual prompts.
    \item \textbf{CXRMate}~\cite{nicolson2024longitudinal}: an encoder–decoder model for CXR report generation trained with reinforcement learning using a semantic similarity reward based on CXR-BERT~\cite{boecking2022making} embeddings.
\end{itemize}

\subsection{Evaluation Metrics}
\label{Metrics}
\noindent
We adopt a comprehensive evaluation protocol, combining quantitative metrics, factual correctness checks, and a Visual Turing Test.
The first evaluates statistical similarity, while the others assesses the clinical plausibility and alignment of the generated outputs.

\subsubsection{Quantitative Metrics}
\noindent
To evaluate generation quality, we use two standard metrics: Fréchet Inception Distance (FID) for images, and Bilingual Evaluation Understudy (BLEU) for text.
\\
\\
FID~\cite{heusel2017gans} measures the dissimilarity between real and synthetic samples in the feature space of a pretrained backbone, ranging in the interval $[0, +\infty)$ with lower values indicating greater similarity.
While standard implementations use Inception-V3 trained on ImageNet~\cite{russakovsky2015imagenet}, we follow best practices in medical imaging~\cite{tronchin2021evaluating} and compute FID using another two backbones, i.e., the XRV-Densenet~\cite{cohen2022torchxrayvision}, an in-domain classifier trained on a large collection of CXRs from different datasets, and XRV-Mimic densenet~\cite{cohen2022torchxrayvision}, specifically trained for MIMIC-CXR scans classification. 
However, to remain coherent with other works, we decided to report the result obtained using the latter backbone in Appendix, in \tablename~\ref{tab:FID-xrvmimic}.
\\
\\
BLEU~\cite{papineni2002bleu} compares machine-generated text to a set of references by calculating the n-gram overlap between the two.
Following the literature, here we computed the BLEU score for a number of n-grams equal to {1,2,3,4}.
BLEU-1 and BLEU-2 place greater emphasis on the consistency of the vocabulary used, focusing on single words or word pairs, while BLEU-3 and BLEU-4 provide information about the semantic structure of the reports.

\subsubsection{Factual Correctness}
\label{Factual Correctness}
\noindent
Assessing factual alignment between generated outputs and clinical ground truth is essential to ensure diagnostic validity, as it evaluates whether the model accurately captures medically relevant information.
\\
\\
\textbf{X-rays Classification}: To evaluate whether the models are capable of generating images that accurately reflect the information of the corresponding clinical reports, we classified the generated samples using XRV-DenseNet~\cite{cohen2022torchxrayvision}.
Since such a classifier is trained only on a subset of pathologies, we computed the AUC and F1-Scores for the following diseases: Atelectasis (Atl.), Cardiomegaly (Cmgl.), Consolidation (Cnsl.), Edema (Edm.), Enlarged Cardiomediastinum (Enl.), Lung Lesion (Les.), Lung Opacity (Opc.), Pleural Effusion (Eff.), Pneumonia (Pnm.) and Pneumothorax (Ptx.), along with micro, macro, and weighted averages.
The micro average aggregates contributions from all classes to provide an overall measure, the macro average computes the metric independently for each class and averages them, and the weighted average adjusts the macro average by accounting for the number of samples per class.
However, because not all scans in the MIMIC-CXR have a defined label for every pathology, we computed the performance for each class only when a ground truth was available, \tablename~\ref{tab:PosDist} report the number of samples for every pathology in our test set.
\begin{table}[h]
\centering
\resizebox{0.6\columnwidth}{!}{
\begin{tabular}{cccc}
\toprule
\textbf{Pathology} & \textbf{\# of Samples} & \textbf{\# of Negatives} & \textbf{\# of Positives} \\ \midrule
Atl. & 10561 & 9123 & 1438 \\
Cgml. & 10305 & 9159 & 1146 \\
Cnsl. & 9911 & 9657 & 254 \\
Edm. & 10230 & 9456 & 774 \\
Enl. & 9215 & 9132 & 83 \\
Les. & 9476 & 9134 & 342 \\
Opc. & 11136 & 9156 & 1980 \\
Eff. & 10558 & 9300 & 1258 \\
Pnm. & 10454 & 9601 & 853 \\
Ptx. & 9337 & 9253 & 84 \\
\bottomrule
\end{tabular}
}
\caption{Number of samples for every pathology for the X-rays classification task.}
\label{tab:PosDist}
\end{table}
\\
\\
\textbf{Report Classification}: For report classification, we leveraged CheXpert-Labeler~\cite{irvin2019chexpert}, a rule-based natural language processing tool that reads a text report and extracts whether it mentions the presence or absence of significant radiologic findings.
Since a rule-based classifier is used, it is not possible to compute the AUC; instead, we reported the F1-Score for the same subset of disease previously introduced along with No Finding (No F.) class.
To remain consistent with the previous setup, we also reported the micro, macro, and weighted averages for the F1-Score.
This task quantifies the ability of the model to generate reports that align with the medical conditions seen in the X-ray images, ensuring that the synthetic reports accurately reflect the diagnostic information provided by the images.
\\
\\
In addition to classification-based evaluation, we also report two commonly adopted metrics for radiology report generation: 
\\
\\
\textbf{RadGraph-F1}~\cite{delbrouck2022improving} evaluates reports by parsing them into a structured graph of radiology-specific entities (e.g., anatomical locations, observations) and their relationships (e.g., "opacity located in left lung"). 
The F1-Score reflects the overlap between ground-truth and generated graphs in terms of both node and edge accuracy, capturing whether the report conveys the correct clinical information and context.
\\
\\
\textbf{RaTE Score}~\cite{zhao2024ratescore} provides a composite measure of factuality and consistency. 
It uses a clinically informed extraction pipeline to identify key findings and anatomical references in the report, and compares them with ground-truth annotations while accounting for linguistic variations, negations, and uncertainty. 
The final score reflects both the correctness and the logical consistency of the report.

Together, these metrics offer a comprehensive and clinically grounded assessment of the quality of the generated outputs.

\subsubsection{Visual Turing Test}
\noindent
We performed a qualitative assessment of the data generated by \mbox{XGeM}, through a Visual Turing Test performed by three expert radiologist. 
This evaluation consisted of five independent tasks aimed at comparing synthetic and real medical data, with both X-rays and clinical report being assessed.
Each task was performed through a web-based platform, where experts evaluated the data using a 1-to-5 numeric scale. 
A score of 1 indicated the poorest quality, while a score of 5 represented the highest level of quality and coherence. 
The tasks are:
\begin{itemize}
    \item \textbf{General X-ray Realism:} Experts rated the overall realism of a series of 20 images, which included a mix of real and synthetic X-rays.
    A high score indicates that the synthetic X-rays appear indistinguishable from real clinical X-rays, with accurate anatomical structures and no visible artifacts that could mislead a clinician in a diagnostic setting.
    \item \textbf{General Report Realism:} Experts reviewed 20 clinical reports, both real and synthetic, rating their plausibility and realism.
    A high score in this task implies that the synthetic reports accurately reflect the clinical context, making use of appropriate medical terminology and providing clinically relevant findings that would be consistent with a real report.
    \item \textbf{Report Coherence with Pair of X-rays:} Experts evaluated the consistency between 20 pairs of X-ray images and their associated reports. The images were guaranteed to be real, while the reports were either real or synthetic, presented in random order.
    A high score indicates that the synthetic reports align accurately with the X-ray images, with no contradictions or inconsistencies between the reported findings and the visual evidence.
    \item \textbf{Coherence Between Report and X-ray:} Experts compared 20 pairs of X-ray images, both real and synthetic, in random order, with the real clinical report to assess the plausibility of the X-ray given the report.
    A high score reflects that the synthetic X-rays match the findings described in the report, indicating that the generative model correctly understood and translated the clinical context from the report into the visual representation.
    \item \textbf{Coherence Between X-ray Pairs:} Experts assessed the consistency between 20 pairs of frontal and lateral X-rays. One of the view was guaranteed to be real, while the other were either real or synthetic, presented in random order.
    A high score here indicates that the synthetic view accurately represents the same clinical findings as the real view, demonstrating proper anatomical alignment and no conflicting features across different perspectives.
\end{itemize}

\section{Results and Discussion}
\noindent
This section provides an in-depth assessment of \mbox{XGeM}'s performance through a comprehensive set of quantitative and qualitative evaluations, including visual examples.

\subsection{X-ray Generation}
\begin{figure}[htb]
\centering
\includegraphics[width=0.85\textwidth]{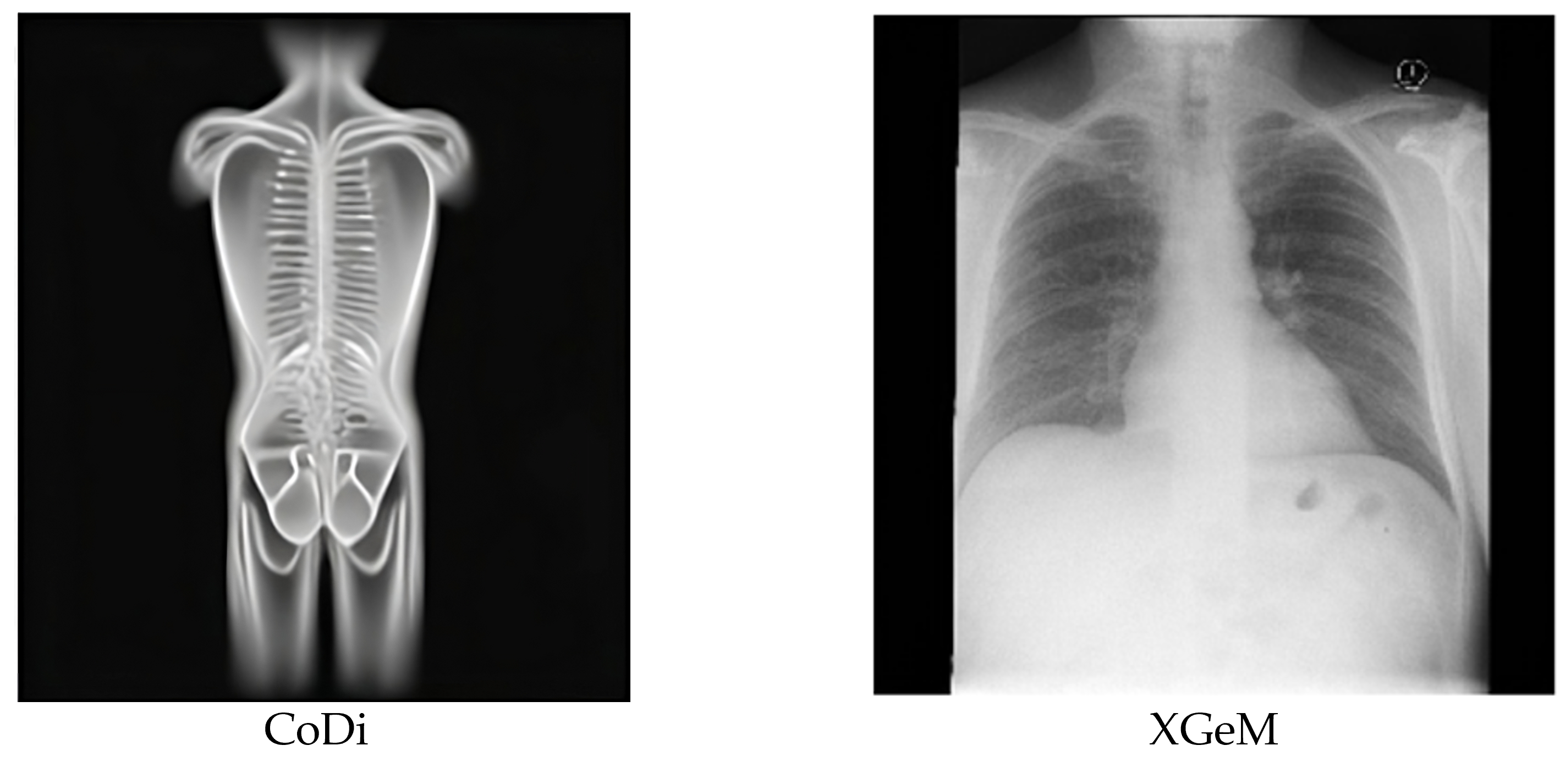}
\caption{Generation comparison between CoDi and XGeM using the same textual prompt, i.e., ``Chest X-ray presents no acute cardiopulmonary process''.}\label{fig:comparison}
\end{figure}

\begin{figure*}[htb]
\centering
\includegraphics[width=0.9\textwidth]{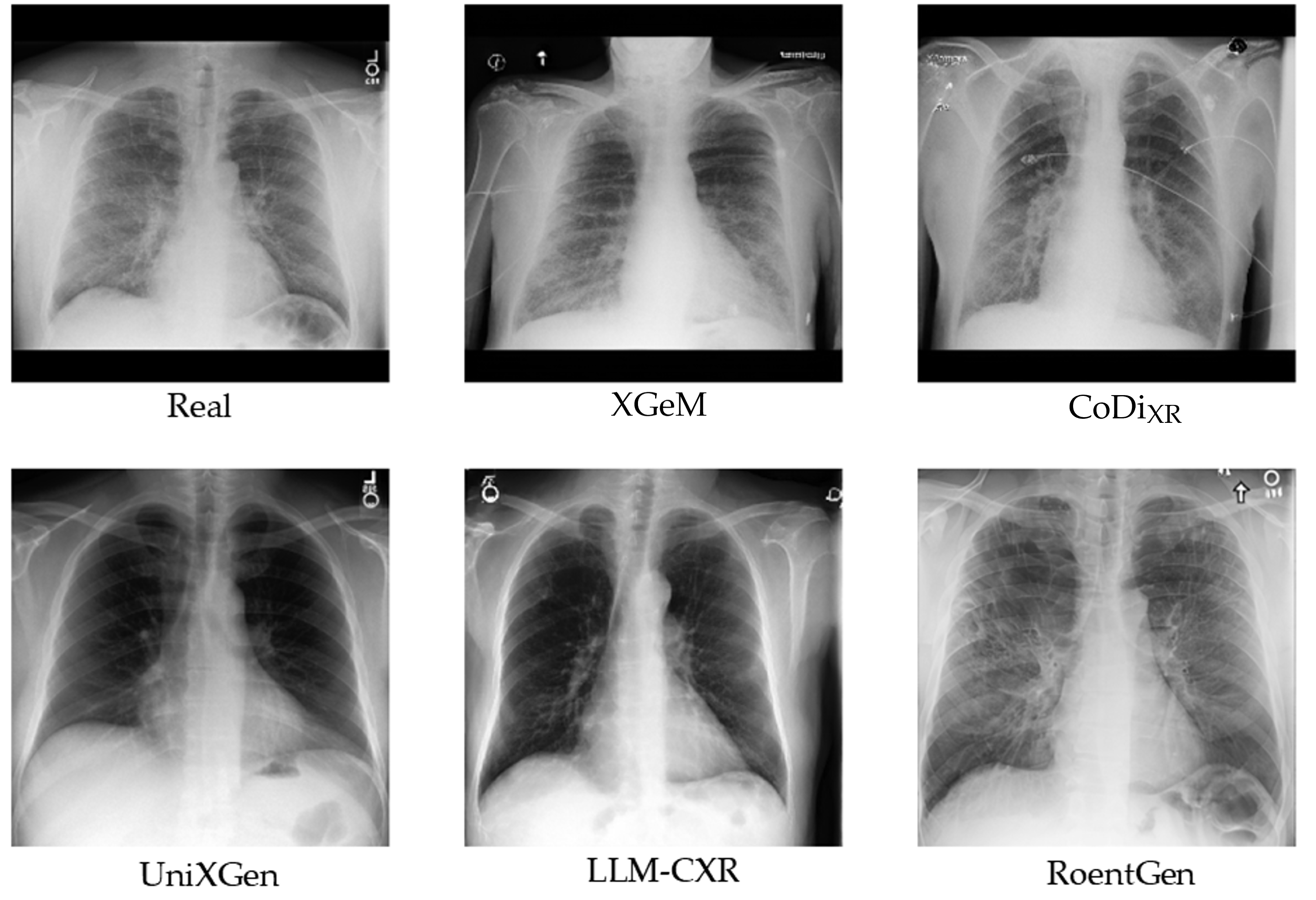}
\caption{Frontal Synthetic Samples generated by different baselines with the same input prompt, i.e., ``No acute cardiopulmonary process''.\label{fig:frontal}}
\end{figure*}

\begin{figure*}[htb]
\centering
\includegraphics[width=0.9\textwidth]{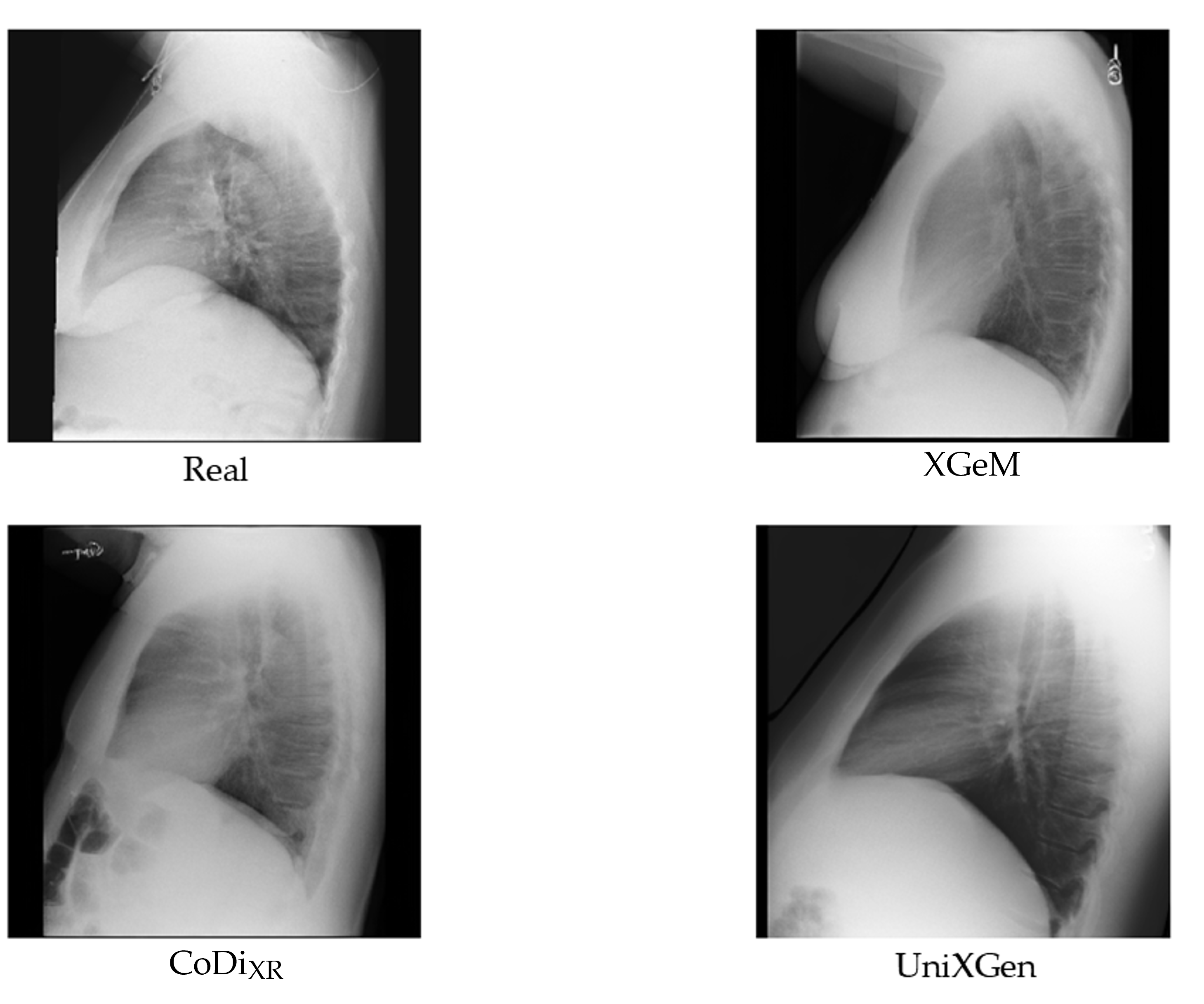}
\caption{Lateral Synthetic Samples generated by different baselines with the same input prompt, i.e., ``No acute cardiopulmonary process''.\label{fig:lateral}}
\end{figure*}

\noindent
\tablename~\ref{tab:FID} reports the FID scores on the test set for generating both frontal (F) and lateral (L) chest X-rays.
The first column lists the models used for generation, while the remaining columns report performance across various generation settings: from clinical report to frontal or lateral CXR (T$\rightarrow$F, T$\rightarrow$L), from one view to the other (L$\rightarrow$F, F$\rightarrow$L), and from a combination of report and image to the missing view (L+T$\rightarrow$F, F+T$\rightarrow$L).
\begin{table}[t]
\centering
\resizebox{\textwidth}{!}{
\begin{tabular}{lcccccccccccc}
\toprule
\textbf{Model} & \multicolumn{2}{c}{\textbf{T$\rightarrow$F}} & \multicolumn{2}{c}{\textbf{L$\rightarrow$F}} & \multicolumn{2}{c}{\textbf{L$+$T$\rightarrow$F}} & \multicolumn{2}{c}{\textbf{T$\rightarrow$L}} & \multicolumn{2}{c}{\textbf{F$\rightarrow$L}} & \multicolumn{2}{c}{\textbf{F$+$T$\rightarrow$L}} \\
\cmidrule(lr){2-3} \cmidrule(lr){4-5} \cmidrule(lr){6-7} \cmidrule(lr){8-9} \cmidrule(lr){10-11} \cmidrule(lr){12-13}
& \textbf{v3 ↓} & \textbf{XRV ↓} & \textbf{v3 ↓} & \textbf{XRV ↓} & \textbf{v3 ↓} & \textbf{XRV ↓} & \textbf{v3 ↓} & \textbf{XRV ↓} & \textbf{v3 ↓} & \textbf{XRV ↓} & \textbf{v3 ↓} & \textbf{XRV ↓} \\
\midrule
\multirow{6}{*}{\begin{tabular}[c]{@{}l@{}} RoentGen \\ UniXGen \\ LLM-CXR \\ CoDi \\ $\text{CoDi}_{\mathit{{XR}}}$ \\ XGeM \end{tabular}}
& 102.77 & 5.60 & - & - & - & - & - & - & - & - & - & - \\
& 81.75 & 7.28 & - & - & 86.21 & 7.63 & 128.96 & 9.76 & - & - & 133.38 & 10.20 \\
& 71.91 & 6.83 & - & - & - & - & - & - & - & - & - & - \\
& 541.44 & 107.23 & 520.02 & 83.12 & 539.35 & 107.17 & 522.00 & 83.01 & 540.66 & 105.63 & 525.80 & 82.52 \\
& \textbf{10.56} & \textbf{0.86} & 34.89 & 3.31 & 22.63 & 1.90 & \textbf{13.90} & \textbf{0.84} & 43.24 & 4.95 & 23.12 & 1.99 \\
& 10.67 & 0.93 & \textbf{12.04} & \textbf{0.48} & \textbf{11.51} & \textbf{0.43} & 14.00 & 0.96 & \textbf{13.75} & \textbf{0.48} & \textbf{11.97} & \textbf{0.34} \\
\bottomrule
\end{tabular}
}
\captionsetup{justification=raggedright, singlelinecheck=false}
\caption{FID score for X-ray generation, with lower values indicating greater similarity. XRV and v3 refers to the two backbones used to compute the score, respectively XRV-Densenet and Inception-v3. 
The “-” symbol indicates that the respective models are not capable of performing the specified generation task}
\label{tab:FID}
\end{table}
Results in \tablename~\ref{tab:FID} demonstrate that XGeM consistently outperforms all competitors in X-ray generation.
Notably, the excessively high FID scores observed for CoDi underscore the importance of fine-tuning in adapting LDMs to the medical imaging domain.
This observation is further supported by \figurename~\ref{fig:comparison}, which illustrates a comparison between samples generated by CoDi and XGeM using the same textual prompt. While XGeM produces a plausible radiograph, CoDi fails to generate an image that resembles a chest X-ray.
\\
In addition, the effectiveness of the Multi-Prompt training strategy is evident, as \mbox{XGeM} consistently outperforms $\text{CoDi}_{\mathit{{XR}}}$ in four configurations (L$\rightarrow$F, L$+$T$\rightarrow$F, F$\rightarrow$L, F$+$T$\rightarrow$L). This highlights the benefit of merging heterogeneous inputs within a shared latent space to enhance generation performance.
However, since the Multi-Prompt training strategy balances multiple generation tasks during optimization, each specific task is only observed a fraction of the time, approximately one-third in our configuration.
As a result, performance slightly drops when generation relies solely on textual information, not due to a limitation of the model architecture, but rather to the reduced exposure to these specific mappings during training.
Nonetheless, this degradation is marginal compared to the substantial gains achieved in multi-modal configurations, where complementary inputs can be jointly exploited to improve generation.
Figures~\ref{fig:frontal} and~\ref{fig:lateral} display representative synthetic images generated by the different models.
Although visually similar at first glance, the FID scores reveal that XGeM’s outputs exhibit greater fidelity to real X-rays, further supporting the effectiveness of our approach.
\\
To further evaluate the factual correctness of XGeM’s generation, we conducted a pathology classification task based on the synthesized X-rays (Section~\ref{Factual Correctness}).
Specifically, each synthetic image was evaluated using the pre-trained XRV classifier~\cite{cohen2022torchxrayvision}, enabling assessment of clinically relevant features.
The results are presented in Table~\ref{tab:CXR_classification}.
\begin{table*}[ht]
\centering
\begin{adjustbox}{width=\textwidth}
\addtolength{\tabcolsep}{-4pt}  
{\scriptsize 
\begin{tabular}{ll|cccccccccc|ccc}
\toprule
\textbf{Task} & \textbf{Model} & \textbf{Atl.} & \textbf{Cmgl.} & \textbf{Cnsl.} & \textbf{Edm.} & \textbf{Enl.} & \textbf{Les.} & \textbf{Opc.} & \textbf{Eff.} & \textbf{Pnm.} & \textbf{Ptx.} & \textbf{Micro} & \textbf{Macro} & \textbf{Weighted} \\ \midrule
- & Real Data & .84 & .91 & .91 & .93 & .81 & .78 & .85 & .95 & .78 & .86 & .87 & .86 & .87 \\
\midrule
\multirow{5}{*}{T$\rightarrow$F} & RoentGen  & .87 & .93 & .82 & .80 & .50 & .59 & .68 & .94 & .48 & .58 & .78 & .72 & .78 \\
& UniXGen & .75 & .78 & .69 & .81 & .67 & .67 & .69 & .76 & .66 & .66 & .74 & .71 & .71 \\
& LLM-CXR & .89 & .92 & .87 & .96 & .80 & .78 & .85 & .95 & .80 & .85 & .89 & .87 & .87 \\
& $\text{CoDi}_{\mathit{{XR}}}$ & .91 & .95 & .93 & .96 & .84 & .82 & .91 & .97 & .84 & .88 & .90 & .90 & .91 \\ 
& XGeM & .86 & .95 & .91 & .94 & .82 & .75 & .86 & .97 & \textbf{.86} & .85 & .88 & .89 & .90 \\
\midrule
\multirow{2}{*}{L$\rightarrow$F} & $\text{CoDi}_{\mathit{{XR}}}$ & .84 & .88 & .90 & .91 & .81 & .76 & .85 & .95 & .76 & .82 & .83 & .85 & .90 \\
& XGeM & .86 & .92 & .93 & .94 & .83 & .80 & .89 & .97 & .80 & .90 & .87 & .88 & .90 \\
\midrule
\multirow{3}{*}{T$+$L$\rightarrow$F} & UniXGen  & .75 & .79 & .65 & .81 & .66 & .63 & .68 & .77 & .63 & .70 & .73 & .71 & .71 \\
& $\text{CoDi}_{\mathit{{XR}}}$ & .89 & .91 & .92 & .91 & .82 & .80 & .86 & .97 & .77 & .85 & .88 & .89 & .90 \\
& XGeM & \textbf{.92} & \textbf{.96} & \textbf{.95} & \textbf{.97} & \textbf{.87} & \textbf{.85} & \textbf{.92} & \textbf{.98} & .85 & \textbf{.93} & \textbf{.91} & \textbf{.92} & \textbf{.92} \\
\bottomrule
\end{tabular}
}
\end{adjustbox}
\vspace*{0.5cm}

\begin{adjustbox}{width=\textwidth}
\addtolength{\tabcolsep}{-4pt}  
{\scriptsize 
\begin{tabular}{ll|cccccccccc|ccc}
\toprule
\textbf{Task} & \textbf{Model} & \textbf{Atl.} & \textbf{Cmgl.} & \textbf{Cnsl.} & \textbf{Edm.} & \textbf{Enl.} & \textbf{Les.} & \textbf{Opc.} & \textbf{Eff.} & \textbf{Pnm.} & \textbf{Ptx.} & \textbf{Micro} & \textbf{Macro} & \textbf{Weighted} \\ \midrule
- & Real Data & .49 & .49 & .35 & .60 & .09 & .12 & .60 & .73 & .35 & .21 & .49 & .40 & .53 \\
\midrule
\multirow{5}{*}{T$\rightarrow$F} & RoentGen  & .31 & .40 & .00 & .25 & .00 & .02 & .40 & .74 & .04 & .04 & .36 & .25 & .43 \\
& UniXGen & .25 & .31 & .05 & .29 & .06 & .05 & .37 & .41 & .00 & .04 & .24 & .17 & .34 \\
& LLM-CXR & .51 & .53 & .19 & .60 & .08 & .07 & .60 & .74 & .38 & .16 & .47 & .39 & .51 \\
& $\text{CoDi}_{\mathit{{XR}}}$ & \textbf{.60} & \textbf{.60} & .33 & .66 & .10 & .05 & .64 & .78 & .38 & .17 & .53 & .43 & .59 \\
& XGeM & .51 & \textbf{.60} & .25 & .59 & .08 & .11 & .60 & .78 & .33 & .12 & .50 & .41 & .54 \\
\midrule
\multirow{2}{*}{L$\rightarrow$F} & $\text{CoDi}_{\mathit{{XR}}}$ & .51 & .45 & .32 & .42 & .06 & .07 & .59 & .73 & .27 & .10 & .43 & .35 & .49 \\
& XGeM & .51 & .46 & .39 & .57 & .09 & .13 & .64 & .79 & .36 & .17 & .50 & .41 & .54 \\
\midrule
\multirow{3}{*}{T$+$L$\rightarrow$F} & UniXGen  & .20 & .31 & .06 & .27 & .06 & .07 & .34 & .43 & .04 & .07 & .24 & .19 & .33 \\
& $\text{CoDi}_{\mathit{{XR}}}$ & .54 & .55 & .32 & .64 & .09 & .06 & .63 & .75 & .38 & .17 & .53 & .44 & .58 \\
& XGeM & .58 & .59 & \textbf{.39} & \textbf{.68} & \textbf{.13} & \textbf{.18} & \textbf{.68} & \textbf{.81} & \textbf{.40} & \textbf{.26} & \textbf{.57} & \textbf{.47} & \textbf{.60} \\
\bottomrule
\end{tabular}
}
\end{adjustbox}
\caption{CXR classification performance evaluated using XRV-DenseNet. Results are reported in terms of AUC and F1-score across multiple pathology labels. Bold values indicate the best performance for each column.}
\label{tab:CXR_classification}
\end{table*}
XGeM again achieves the highest AUC and F1-Scores, confirming its capacity to faithfully capture and represent clinically relevant features from the input prompts.
While some F1-Scores remain low, likely due to the limited number of positive samples for certain pathologies, as shown in \tablename~\ref{tab:PosDist}, XGeM still outperforms all baselines.
Interestingly, synthetic images outperform real data in some classification settings.
We hypothesize that XGeM’s strong encoding of disease-specific patterns results in more prototypical samples, which are easier for the classifier to recognize.
This hypothesis is reinforced by the results of the Visual Turing Test (Section~\ref{VTTsection}), where synthetic samples were consistently rated as clinically realistic.
These findings suggest that XGeM not only generates visually plausible X-rays, but also preserves key clinical attributes present in real data.
However, this behavior may also reflect a tendency to emphasize common disease patterns, potentially limiting the model's ability to represent subtle or ambiguous cases, those that are also harder to detect by the classification backbone.

\subsection{Report Generation}
\noindent
Table~\ref{BLEU} reports BLEU scores on the test set for report generation under three input conditions: frontal CXR only (F$\rightarrow$T), lateral CXR only (L$\rightarrow$T), and both views combined (F$+$L$\rightarrow$T).
\begin{table*}[ht]
\centering
\begin{adjustbox}{width=\textwidth}
\small
\setlength{\tabcolsep}{5pt}
\begin{tabular}{lcccccccccccc}
\hline
\multirow{2}{*}{Methods} & \multicolumn{4}{c}{\begin{tabular}[c]{@{}c@{}}F$\rightarrow$T\end{tabular}}                        
& \multicolumn{4}{c}{L$\rightarrow$T} & \multicolumn{4}{c}{F$+$L$\rightarrow$T}
\\ \cmidrule(lr){2-5} \cmidrule(lr){6-9} \cmidrule(lr){10-13} 
& BLEU-1 ↑       & BLEU-2 ↑       & BLEU-3 ↑       & BLEU-4 ↑      
& BLEU-1 ↑       & BLEU-2 ↑       & BLEU-3 ↑       & BLEU-4 ↑
& BLEU-1 ↑       & BLEU-2 ↑       & BLEU-3 ↑       & BLEU-4 ↑
\\ \hline
CXRMate
& .29          & .21          & .15          & .11 
& .27          & .19          & .14          & .10 
& .29          & .20          & .14          & .11 \\
UniXGen
& .25          & .16          & .12          & .09     
& .26          & .16          & .11          & .07 
& .26          & .17          & .12          & .09 \\
LLM-CXR
& .25          & .15          & .10          & .07     
& -              & -              & -              & - 
& -              & -              & -              & - \\
$\text{CoDi}_{\mathit{{XR}}}$
& .38          & .27          & .22          & .18     
& .38          & .28        & .23       & .18 
& .42          & .32        & .27       & .22 \\
XGeM
& \textbf{.41} & \textbf{.30} & \textbf{.25} & \textbf{.20} 
& \textbf{.42} & \textbf{.32} & \textbf{.26} & \textbf{.21} 
& \textbf{.44} & \textbf{.34} & \textbf{.29} & \textbf{.24} \\
\bottomrule
\end{tabular}
\end{adjustbox}
\captionsetup{justification=raggedright, singlelinecheck=false}
\caption{BLEU Score for Report Generation, with higher values indicating greater similarity. The ``-'' symbol indicates that the respective models are not capable of performing the specified generation task. Bold values indicate the best performance for each column.}
\label{BLEU}
\end{table*}
Higher BLEU scores indicate greater alignment with the reference reports, providing a measure of both lexical and structural similarity across methods.
XGeM consistently outperforms all baselines across BLEU-1 to BLEU-4, confirming its superiority in both word choice and sentence structure.
These results suggest that XGeM not only adopts appropriate clinical terminology, reflected in its BLEU-1 and BLEU-2 scores, but also constructs sentences with realistic syntactic structure, as evidenced by its BLEU-3 and BLEU-4 performance.
This reflects a high degree of linguistic coherence and fidelity in replicating both the semantics and style of authentic medical reports.
We further assessed the model’s consistency by computing BLEU scores for reports generated from multiple scans belonging to the same clinical study.
This evaluation measures the model's ability to produce coherent and consistent reports across different views of the same patient case.
The objective of this evaluation is to verify not only the quality of single-report generation, but also the model’s robustness in maintaining diagnostic consistency across related inputs.
Numerical results are provided in Appendix, in Table~\ref{tab:bleu_study}.
\\
\\
Similar to the synthetic X-ray evaluation, we performed a classification task on the generated reports to assess whether they accurately reflect the clinical findings provided as input to XGeM.
As shown in Table~\ref{tab: Report-Clf}, XGeM again achieves superior classification performance compared to all baselines.
It obtains the highest F1-scores in nearly all categories, including Micro, Macro, and Weighted averages across all input settings.
These results indicate that XGeM not only generates realistic reports but also accurately encodes the clinical conditions specified in the input.
Notably, performance peaks when both views are used as input (F$+$L$\rightarrow$T), suggesting that integrating multiple perspectives enables the model to form a more complete clinical picture.

\begin{table*}[h]
\begin{adjustbox}{width=\textwidth}
\addtolength{\tabcolsep}{-4pt}  
\begin{tabular}{ll|ccccccccccc|ccc}
\toprule
\textbf{Task} & \textbf{Model} & \textbf{Atl.} & \textbf{Cmgl.} & \textbf{Cnsl.} & \textbf{Edm.}  & \textbf{Enl.} & \textbf{Les.} & \textbf{Opc.} & \textbf{Eff.} & \textbf{Pnm.} & \textbf{Ptx.} & \textbf{No F.} & \textbf{Micro} & \textbf{Macro} & \textbf{Weighted} \\ \midrule
\multirow{5}{*}{F$\rightarrow$T} 
& CXRMate & .33 & .34 & .21 & .58 & .02 & .12 & .18 & .65 & .29 & .01 & .82 & .62 & .35 & .57 \\
& UniXGen  & .18 & .17 & .03 & .28 & .01 & .01 & .01 & .13 & .10 & .03 & .74 & .49 & .15 & .42 \\
& LLM-CXR & .25 & .23 & .07 & .40 & .02 & .03 & .21 & .36 & .24 & .02 & .74 & .46 & .20 & .44 \\
& $\text{CoDi}_{\mathit{{XR}}}$ & .56 & .56 & .15 & .67 & .10 & .26 & .47 & .71 & .52 & .37 & .88 & .67 & .44 & .67 \\
& XGeM & \textbf{.62} & \textbf{.61} & .14 & \textbf{.70} & \textbf{.15} & .27 & .52 & \textbf{.79} & \textbf{.59} & .34 & .90 & .71 & \textbf{.46} & \textbf{.73} \\
\midrule
\multirow{4}{*}{L$\rightarrow$T}
& CXRMate & .35 & .25 & .11 & .47 & .02 & .01 & .29 & .73 & .27 & .01 & .81 & .62 & .23 & .56 \\
& UniXGen  & .23 & .23 & .05 & .36 & .04 & .02 & .20 & .36 & .16 & .05 & .73 & .45 & .19 & .43 \\
& $\text{CoDi}_{\mathit{{XR}}}$ & .54 & .58 & .14 & .65 & .07 & .23 & .45 & .75 & .51 & .30 & .89 & .68 & .41 & .68 \\
& XGeM & \textbf{.62} & \textbf{.61} & .16 & \textbf{.70} & .09 & .26 & .51 & .78 & .57 & .32 & \textbf{.91} & .71 & .45 & \textbf{.73} \\
\midrule
\multirow{4}{*}{F$+$L$\rightarrow$T} 
& CXRMate & .38 & .40 & .17 & .55 & .02 & .10 & .22 & .67 & .31 & .22 & .82 & .62 & .39 & .58 \\
& UniXGen & .22 & .16 & .06 & .30 & .01 & .00 & .17 & .33 & .17 & .02 & .75 & .49 & .16 & .45 \\
& $\text{CoDi}_{\mathit{{XR}}}$ & .61 & .60 & \textbf{.17} & \textbf{.70} & .09 & \textbf{.31} & .51 & .78 & .58 & .27 & .90 & .72 & .45 & .72 \\
& XGeM & \textbf{.62} & \textbf{.61} & \textbf{.17} & \textbf{.70} & .12 & .27 & \textbf{.53} & \textbf{.79} & .58 & \textbf{.37} & \textbf{.91} & \textbf{.73} & \textbf{.46} & \textbf{.73} \\
\bottomrule
\end{tabular}
\end{adjustbox}
\caption{Report classification performance evaluated using CheXPert-Labeler. Results are reported in terms of F1-score across multiple pathology labels. Bold values indicate the best performance for each column.}
\label{tab: Report-Clf}
\end{table*}
In addition, we evaluated the quality of the generated reports using two specialized metrics designed for radiology report generation: RadGraph-F1 and RaTE Score.
The results in Table~\ref{tab:radgraph_rate} show that XGeM outperforms all competitors for both metrics. 
Again, a consistent improvement is observed when both frontal and lateral views are provided as input, confirming the advantage of multi-view conditioning in capturing richer clinical semantics and improving the structural and relational accuracy of the generated reports.

\begin{table*}[ht]
\centering
\begin{adjustbox}{width=\textwidth}
\small
\setlength{\tabcolsep}{5pt}
\begin{tabular}{lcccccccccccc}
\toprule
\multirow{2}{*}{Methods} & \multicolumn{2}{c}{F$\rightarrow$T}                        
& \multicolumn{2}{c}{L$\rightarrow$T} 
& \multicolumn{2}{c}{F$+$L$\rightarrow$T} \\
\cmidrule(lr){2-3} \cmidrule(lr){4-5} \cmidrule(lr){6-7} 
& RadGraph-F1 ↑ & RaTE ↑      
& RadGraph-F1 ↑ & RaTE ↑      
& RadGraph-F1 ↑ & RaTE ↑ \\
\midrule
CXRMate   & .40 & .51 & .37 & .49 & .39 & .51 \\
UniXGen   & .31 & .47 & .36 & .47 & .36 & .47 \\
LLM-CXR   & .31 & .45 & -    & -    & -    & -    \\
$\text{CoDi}_{\mathit{{XR}}}$   & .49 & .67 & .48 & .67 & .50 & .67 \\
XGeM & \textbf{.56} & \textbf{.72} & \textbf{.54} & \textbf{.71} & \textbf{.58} & \textbf{.72} \\
\bottomrule
\end{tabular}
\end{adjustbox}
\captionsetup{justification=raggedright, singlelinecheck=false}
\caption{Evaluation of report generation using RadGraph and RaTE Score across different input configurations. Higher values indicate better alignment with the ground truth. The “-” symbol denotes unsupported generation tasks. Bold values indicate the best performance for each column.}
\label{tab:radgraph_rate}
\end{table*}

\subsection{Multi-Output Generation}
\begin{figure}[ht]
\vspace{0.5cm}
\centering
\includegraphics[width=105mm]{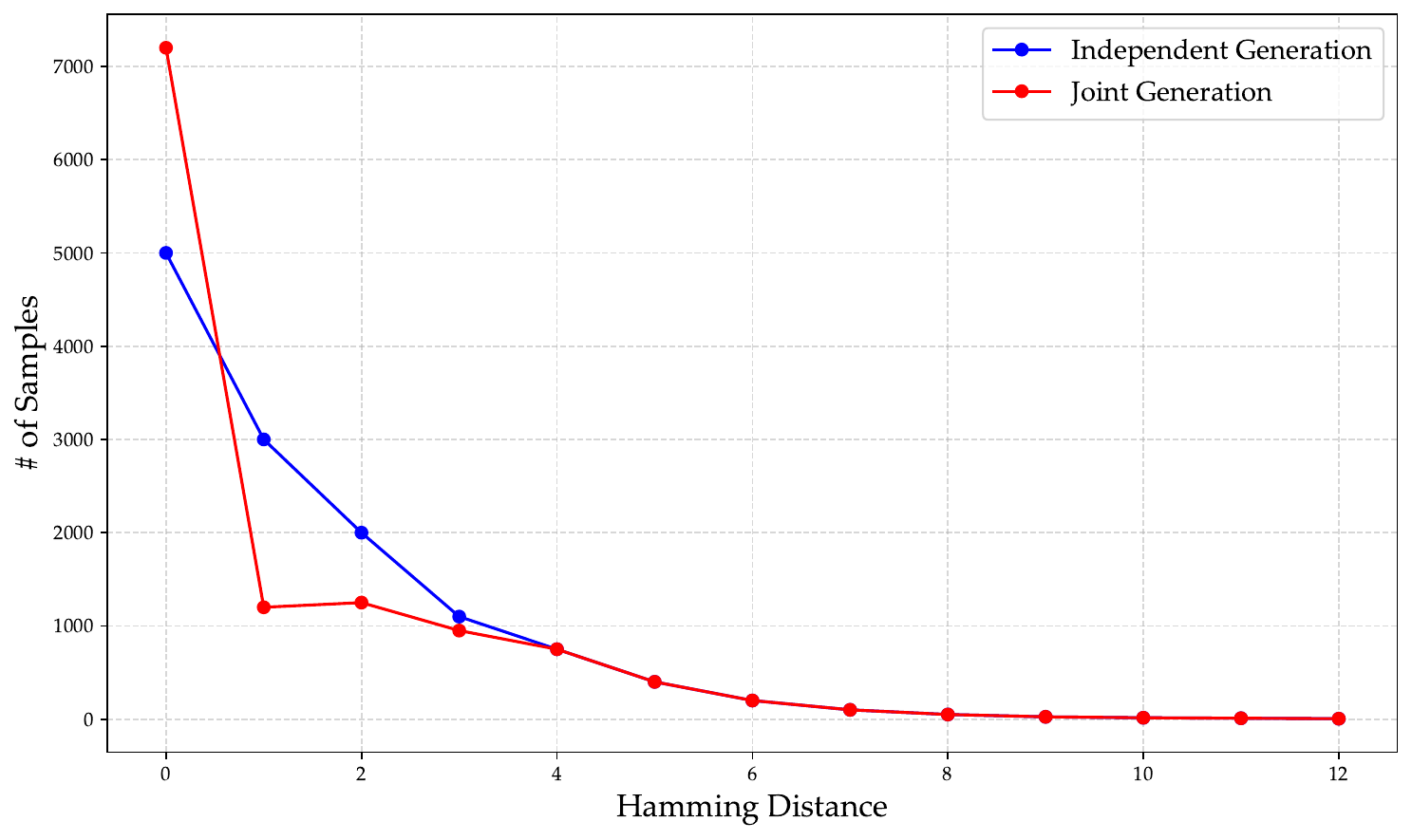}
\caption{Distribution of Hamming distances between clinical labels predicted from generated frontal X-rays and their corresponding reports. We compare two generation strategies: joint generation (red curve) and independent generation (blue curve). Each curve represents the frequency of mismatches (y-axis) for a given Hamming distance (x-axis), i.e., the number of clinical labels that differ between the two modalities.}
\label{fig:hamming}
\end{figure}

\noindent
Leveraging the multi-output training procedure detailed in Section~\ref{Multi-Output Generation}, XGeM learns to generate multiple modalities simultaneously while preserving their semantic consistency.
As no standard metric exists to evaluate the coherence of simultaneously generated modalities, we adopt the cosine similarity between their latent representations $M_i$ and $M_j$~\cite{tang2023anytoany}:
\begin{equation}
    \cos \left(P_i(M_i), P_j(M_j)\right)
\end{equation}
where $P_i$ and $P_j$ denote the prompt encoders that map the respective modalities into the shared latent space.
Intuitively, the closer is the metric to 1, the higher the alignment between the two modalities.
Since none of the baseline models supports multi-output generation, we compare XGeM's joint generation against a sequential approach, where each modality is independently synthesized.
As reported in Table~\ref{tab: Similarity-Scores}, joint generation consistently yields higher similarity scores, indicating better alignment between modalities.
\begin{table}[h]
\centering
\resizebox{0.6\columnwidth}{!}{
\begin{tabular}{c|cc}
\toprule
\textbf{Input} & \textbf{Independent Gen.} & \textbf{Multi-Output Gen.} \\ \midrule
T$\rightarrow$F+L & .61 & \textbf{.65} \\
L$\rightarrow$F+T & .12 & \textbf{.22} \\
F$\rightarrow$L+T & .19 & \textbf{.21} \\
\bottomrule
\end{tabular}
}
\caption{Cosine similarity between latent representations of jointly generated modality pairs under different input configurations. We compare XGeM's Multi-Output generation against a baseline strategy where each output modality is generated independently using the same conditioning input. Higher cosine similarity values indicate greater alignment between the generated modalities within the shared latent space, and thus stronger semantic coherence.}
\label{tab: Similarity-Scores}
\end{table}
To investigate whether higher similarity also corresponds to improved clinical consistency, we compared the coherence between frontal X-rays and reports generated via independent (L$\rightarrow$F $+$ L$\rightarrow$T) versus joint (L$\rightarrow$F$+$T) generation.
Specifically, we aimed to assess whether the two generated modalities were aligned in terms of clinical content. 
To this end, we applied the pretrained classifiers to both the generated images and reports, obtaining two sets of clinical labels for each sample.
We then computed the Hamming Distance~\cite{hamming1950error} to quantify the misalignment between these label sets.
In this context, a smaller Hamming Distance implies stronger agreement between modalities in terms of clinical content.
The distribution of Hamming distances is shown in \figurename~\ref{fig:hamming}, with the x-axis denoting the number of mismatched labels and the y-axis indicating their frequency.
As illustrated, the red curve (joint generation) peaks at lower distances than the blue curve (independent generation), indicating improved alignment.
Specifically, a large proportion of the samples generated via the joint approach exhibit a Hamming Distance equal to zero, meaning the generated modalities are perfectly aligned. 
Furthermore, the lower frequency of mismatches at distances 1 and 2 reinforces the improved clinical coherence of the joint generation.
Beyond distance 2, both curves flatten, indicating that extreme incoherence is uncommon in either setting.
Overall, these results highlight the advantage of XGeM’s Multi-Output generation, which ensures stronger consistency across modalities and better preservation of shared clinical content.  

\subsection{Visual Turing Test}
\label{VTTsection}
\begin{figure*}[htb]
\centering
\includegraphics[width=1\textwidth]{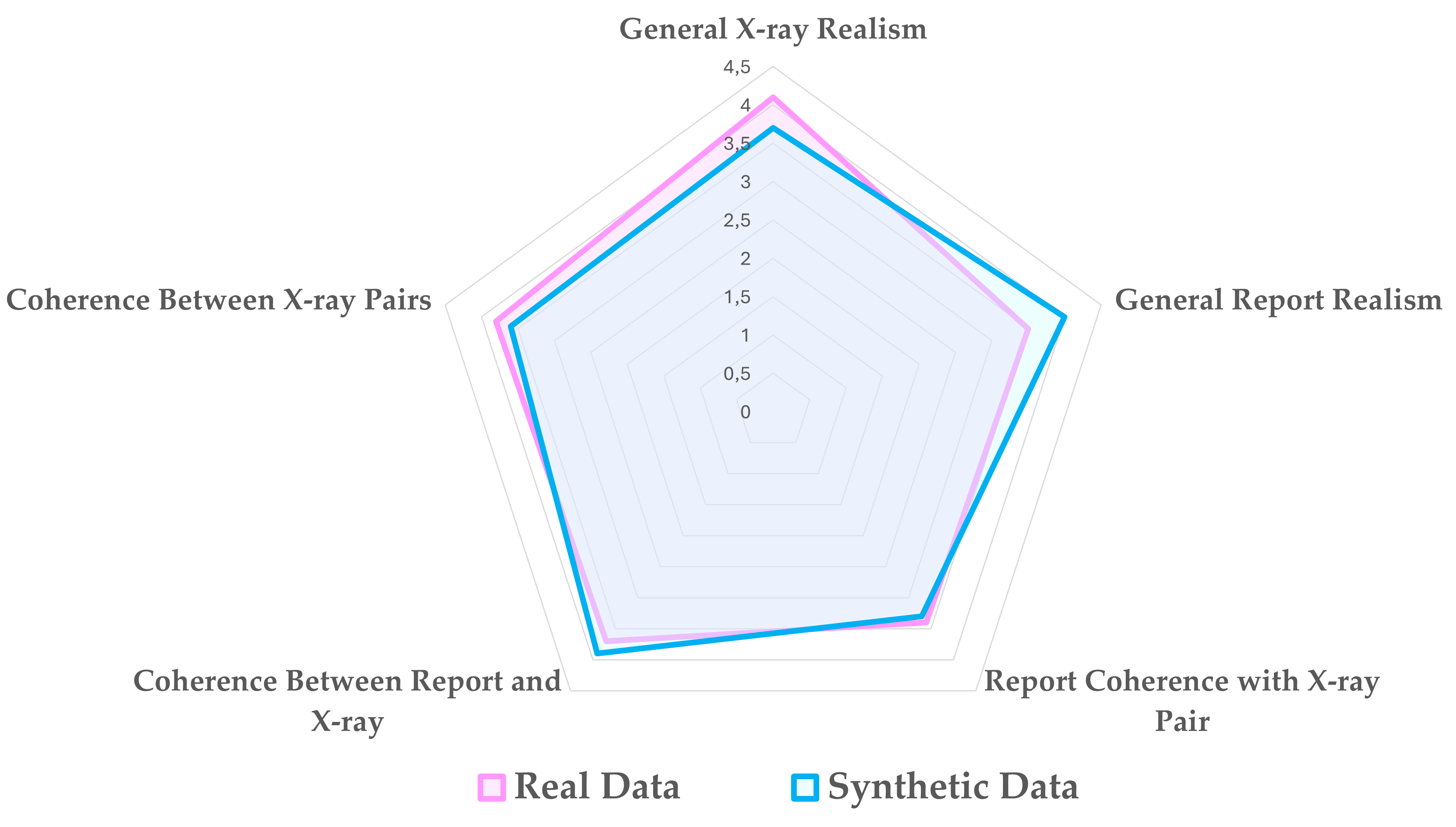}
\caption{Radar chart showing the results of the Visual Turing Test across five dimensions: X-ray realism, Report realism, Report/X-ray coherence, Report/X-ray pair coherence, and X-ray pair consistency. Evaluations were performed by three board-certified radiologists using a 5-point Likert scale, with higher values indicating better performance. \label{fig:radar}}
\end{figure*}

\noindent
To assess the perceptual quality and coherence of XGeM’s synthetic outputs, we conducted a Visual Turing Test involving three board-certified radiologists, each with over 10 years of experience. 
The results, summarized in Figure~\ref{fig:radar}, reflect the average ratings across five evaluation tasks.
For completeness, the numerical scores are reported in Appendix, in \tablename~\ref{tab:VTT_results}.

\paragraph{General X-ray Realism}
\noindent
\\
The average rating for real X-rays was \(\mathbf{4.1}\), demonstrating the radiologists' high level of confidence in recognizing and evaluating authentic clinical images. 
This result highlights the robustness of the evaluation process and the expertise of the participants.  
Synthetic images generated by XGeM scored \(\mathbf{3.7}\), which, while slightly lower, still represents a strong performance, indicating the overall quality of XGeM's outputs. 

\paragraph{General Report Realism}
\noindent
\\
Real clinical reports received an average rating of \(\mathbf{3.5}\), highlighting that even authentic reports are not always perceived as perfect. 
This may be due to occasional discrepancies in interpretation between radiologists or variability in the detail and clarity of the reports, which are often influenced by stylistic differences among clinicians.  
In contrast, synthetic reports generated by XGeM achieved a higher average rating of \(\mathbf{4.0}\). 
This result underscores the model's ability to consistently use accurate medical terminology and construct coherent and contextually appropriate narratives. 

\paragraph{Report Coherence with Pair of X-rays}
\noindent
\\
The average score for real reports was \(\mathbf{3.4}\), while synthetic reports achieved a comparable score of \(\mathbf{3.3}\).
These relatively modest ratings reflect the inherent difficulty of this task, which requires precise alignment between diagnostic findings described in the report and the visual evidence provided by the X-ray pair.
Despite the challenge, XGeM demonstrates the ability to generate diagnostic reports that are closely aligned with those written by radiologists, highlighting the model's capability to produce coherent and contextually relevant reports, even in a highly demanding evaluation scenario.

\paragraph{Coherence Between Report and X-ray}
\noindent
The average score for real X-rays paired with their corresponding reports was \(\mathbf{3.7}\), while synthetic pairs generated by XGeM achieved a slightly higher score of \(\mathbf{3.9}\).
These results highlight the strong capabilities of XGeM in generating X-rays that align closely with the diagnostic context described in the corresponding reports.  
Additionally, while the model's robust understanding of pathologies contributes to producing samples where disease features are prominently and clearly represented, this characteristic might also pose a limitation. 
By generating images with overly evident pathological signs, the synthetic data may fail to capture more subtle or ambiguous cases that often challenge radiologists in real-world clinical practice.

\paragraph{Coherence Between X-ray Pairs}
\noindent
The average score for real X-ray pairs was \(\mathbf{3.8}\), while synthetic pairs scored \(\mathbf{3.6}\).
This suggests that XGeM is able to guarantee anatomical consistency between frontal and lateral views of the same subject, approaching the realism of real X-ray data.

\subsection{About the utility of Synthetic Data in the Medical Field}
\noindent
To further explore the practical value of XGeM in a clinical workflow, we assess its ability to generate synthetic data capable of addressing three critical challenges in the medical field: anonymization, class imbalance, and data scarcity.
These challenges frequently arise in real-world clinical scenarios, where privacy regulations, class imbalance, and limited data availability constrain the development of effective and generalizable models.
For this evaluation, we conducted experiments using both frontal and lateral synthetic X-rays. To ensure clarity, we report only the results for frontal views in the main text, while corresponding lateral-view results are included in the Appendix, in \tablename s~\ref{tab:Anonymization-Lat}, \ref{tab:Data-imbalance-lat}, \ref{tab:Data-scarcity-lat}.

\paragraph{Anonymization}
\noindent
Medical data is governed by strict privacy regulations, which limit its availability for research and model training.
Traditional anonymization techniques, while effective at protecting privacy, often compromise data integrity by removing clinically relevant features.
Synthetic data offers a promising alternative by replicating the distribution of real patient data, enabling privacy-preserving training without sacrificing data utility. 
To evaluate XGeM’s ability to address this challenge, we trained two DenseNet-121 classifiers from scratch: one using only synthetic X-rays, and the other using only real images.
Both models were then evaluated on the real test set to assess whether training exclusively on synthetic data leads to a drop in classification performance.
As shown in Table~\ref{tab:Anonymization}, both classifiers achieve comparable results.
While the classifier trained on synthetic data shows a marginal drop in AUROC, this is balanced by a higher F1-score, indicating that synthetic data can serve as a viable substitute for real data without compromising model effectiveness.

\begin{table}[t]
\centering
\resizebox{0.7\columnwidth}{!}{
\begin{tabular}{lcccccc}
\toprule
\textbf{Training Data} & \multicolumn{3}{c}{\textbf{AUROC}} & \multicolumn{3}{c}{\textbf{F1-Score}} \\ 
\cmidrule(lr){2-4} \cmidrule(lr){5-7}
& \textbf{Micro} & \textbf{Macro} & \textbf{Weighted} & \textbf{Micro} & \textbf{Macro} & \textbf{Weighted} \\
\midrule
\multirow{2}{*}{\begin{tabular}[c]{@{}l@{}} Real \\ Synthetic \end{tabular}}       
& .75 & .74 & .74 & .30 & .20 & .21 \\
& .74 & .73 & .73 & .34 & .23 & .23 \\
\bottomrule
\end{tabular}
}
\caption{Classification performance of two DenseNet-121 models trained respectively on real frontal CXR and on a combination of real and synthetic frontal CXR, evaluated on a held-out test set of real X-rays. Metrics include AUROC and F1-score. This experiment simulates an anonymization scenario, assessing whether synthetic data generated by XGeM can effectively substitute real data for model training without compromising performance.}
\label{tab:Anonymization}
\end{table}

\paragraph{Imbalance Learning}
\noindent
Class imbalance is a well-known issue in machine learning, where models tend to favor majority classes, leading to poor sensitivity in detecting rare but clinically significant conditions.
One possibile solution to address this issue involves augmenting minority classes with synthetic data, helping models generalize more effectively on imbalanced datasets.
To assess XGeM's capabilities to tackle this challenge, we focused on a 5-class classification task, selecting the five most represented classes from our dataset, which are listed in \tablename~\ref{tab:Distribution} along with the respective percentage of positive samples. 
This subset was selected to avoid extreme imbalance, which would otherwise require generating an impractical volume of synthetic data.
For this experiment, we generated synthetic samples to build a training split that ensured a balanced distribution of positive samples across all five classes.
\begin{table}[h]
\centering
\resizebox{0.4\columnwidth}{!}{
\begin{tabular}{lcc}
\toprule
\textbf{Condition} & \textbf{Positive (\%)} \\ \midrule
Atelectasis & 15.6\% \\
Cardiomegaly & 13.5\% \\
Consolidation & 2.99\% \\
Edema & 10.4\% \\
Effusion & 14.1\% \\
\bottomrule
\end{tabular}
}
\caption{Percentage of positive samples in the training set for the condition used for Data Imbalance task.}
\label{tab:Distribution}
\end{table}
\\
Subsequently, we trained two DenseNet-121 models from scratch: one on the original imbalanced dataset, and another on an augmented version with balanced synthetic data.
Both models were trained with identical hyperparameters to ensure a controlled and fair comparison.
In Table~\ref{tab:data-imbalance}, we compare the two classifiers' F1-score on the real test set: the results show an improvement for all classes after synthetic data augmentation, indicating that synthetic data can enhance the classifier’s ability to generalize across all conditions, ensuring better overall performance and sensitivity to rare, clinically important cases. 

\begin{table}[t]
\centering
\resizebox{\textwidth}{!}{
\begin{tabular}{lcccccccccc}
\toprule
\textbf{Training Data} & \multicolumn{8}{c}{\textbf{F1-Score}} \\ 
\cmidrule(lr){2-9}
& \textbf{Atl.} & \textbf{Cmgl.} & \textbf{Cnsl.} & \textbf{Edm.} & \textbf{Eff.} & \textbf{Micro} & \textbf{Macro} & \textbf{Weighted} \\
\midrule
Real & .34 & .42 & .17 & .67 & .64 & .50 & .45 & .45 \\
Real $+$ Synthetic & .48 & .48 & .33 & .69 & .72 & .58 & .54 & .54 \\
\bottomrule
\end{tabular}
}
\caption{F1-score performance of DenseNet-121 classifiers trained on an imbalanced dataset of frontal CXR versus a balanced dataset obtained through XGeM–generated synthetic data. Each column reports the per-class F1-score computed on the real held-out test set. This experiment evaluates whether synthetic data can mitigate class imbalance and improve sensitivity to underrepresented but clinically significant conditions.}
\label{tab:data-imbalance}
\end{table}

\paragraph{Data Scarcity}
\noindent
Collecting large, well-annotated medical datasets remains a major bottleneck, particularly for rare diseases or emerging clinical conditions.
Data scarcity limits the development of robust machine learning models, which typically require substantial amounts of diverse and high-quality data to generalize well.
A promising solution is to augment real datasets with synthetic data that preserves the underlying distribution of the real samples, allowing training on a broader sample space.
To investigate whether XGeM can mitigate this issue, we simulate a low-data regime by creating a reduced training subset, corresponding to approximately 15,000 real X-rays.
We then progressively expand this subset by adding synthetic images generated by XGeM.
For each configuration, we train a DenseNet-121 classifier from scratch and evaluate it on the same real test set.
Performance is reported in terms of AUROC and F1-score across micro, macro, and weighted averages.
As shown in Table~\ref{tab:data-scarcity}, augmenting with XGeM's synthetic samples consistently improves classification performance compared to the low-data baseline.
These results confirm that synthetic data can effectively scale dataset size and enhance model performance in scarce-data scenarios.

\begin{table}[t]
\centering
\resizebox{\textwidth}{!}{
\begin{tabular}{ccc|ccc|ccc}
\toprule
\textbf{Real} & \textbf{Synthetic} & \textbf{Total} & \textbf{AUROC (Micro)} & \textbf{AUROC (Macro)} & \textbf{AUROC (Weighted)} & \textbf{F1 (Micro)} & \textbf{F1 (Macro)} & \textbf{F1 (Weighted)} \\
\midrule
15k  & 0     & 15k   & 0.68 & 0.66 & 0.67 & 0.21 & 0.14 & 0.15 \\
15k  & 5k    & 20k   & 0.70 & 0.69 & 0.69 & 0.24 & 0.17 & 0.18 \\
15k  & 15k   & 30k   & 0.72 & 0.71 & 0.71 & 0.27 & 0.19 & 0.20 \\
15k  & 30k   & 45k   & 0.74 & 0.72 & 0.73 & 0.29 & 0.21 & 0.20 \\
15k  & 60k   & 75k   & 0.74 & 0.73 & 0.73 & 0.33 & 0.23 & 0.22 \\
\bottomrule
\end{tabular}
}
\caption{Classification performance of DenseNet-121 models trained on subsets of 15k real frontal CXRs augmented with increasing amounts of synthetic data from XGeM. Metrics are reported on the same real test set.}
\label{tab:data-scarcity}
\end{table}

\section{Conclusions}
\noindent
This work presents XGeM, a novel foundation model specifically designed for multimodal medical data generation, leveraging diffusion models and contrastive learning. 
The results demonstrate that XGeM excels in generating both realistic chest X-rays and high-quality radiology reports, consistently outperforming state-of-the-art models across both quantitative and factual correctness metrics. 
The main novelty introduced by \mbox{XGeM} is the Multi-Prompt Training strategy, which plays a pivotal role in enhancing cross-modal generation performance, helping the model to capture the complex relationships between medical data modalities.
This approach allows XGeM to effectively integrate and align information from multiple sources in a shared latent space, thereby improving both the visual fidelity of the generated images and the clinical accuracy of the corresponding textual reports.
The ability of XGeM to seamlessly fuse diverse inputs ensures a more cohesive representation, facilitating more accurate and context-aware generation of synthetic medical data, as shown by the results in comparison with $\text{CoDi}_{\mathit{{XR}}}$.
The Visual Turing Test results demonstrate that XGeM performs well in generating synthetic medical data that is both realistic and contextually coherent. 
Across all tasks, synthetic data scored consistently between 3.0 and 4.0, indicating a high level of quality and alignment with real clinical data.  
The model's high performance in tasks requiring consistency between different image views and between reports and images further confirms its potential for generating clinically relevant synthetic data.  
Moreover, the results show that the synthetic data generated by XGeM hold significant promise for addressing challenges in medical research and healthcare, such as anonymization, data imbalance, and data scarcity, suggesting that synthetic chest X-rays and radiology reports can offer new avenues for enhancing AI integration in the medical field. 
Future research could explore several promising directions to further extend the capabilities of XGeM. 
One area of exploration involves scaling the modular framework of XGeM to support additional modalities beyond 2D chest X-rays, such as 3D medical images (e.g., CT, MRI) and time series (e.g., ECG, EEG). 
Additionally, there is significant potential to investigate more advanced techniques for merging input embeddings, which could enhance the integration of heterogeneous medical data sources, for a more holistic understanding of patient conditions. 
The inclusion of longitudinal information and temporal dynamics could also play a vital role in temporal data generation, allowing XGeM to model disease progression and treatment responses over time. 
This would be particularly beneficial for chronic conditions where patient data evolve across multiple time points.
Finally, future work could explore the integration of active learning and reinforcement learning paradigms into the training process, enabling XGeM to iteratively refine its generative capabilities by receiving feedback from expert clinicians or interacting with clinical environments. 
This would ensure that the model continues to improve over time, staying up-to-date with the latest medical knowledge and standards of care. 
Ultimately, XGeM represents a significant step forward in the generation of synthetic medical data, and with further research and development, it has the potential to become a key tool in advancing medical AI across a wide range of applications.

\section*{Author Contributions}
\noindent
\textbf{Daniele Molino:} Conceptualization, Data curation, Formal analysis, Investigation, Methodology, Software, Validation, Visualization, Writing – original draft, Writing – review \& editing;
\textbf{Francesco Di Feola:} Conceptualization, Formal analysis, Investigation, Methodology, Supervision, Validation, Writing – original draft, Writing – review \& editing;
\textbf{Eliodoro Faiella:} Validation;
\textbf{Deborah Fazzini:} Validation;
\textbf{Domiziana Santucci:} Validation;
\textbf{Linlin Shen:} Validation, Writing – review \& editing;
\textbf{Valerio Guarrasi:} Conceptualization, Formal analysis, Investigation, Methodology, Project administration, Resources, Supervision, Validation, Writing – review \& editing.
\textbf{Paolo Soda}  Conceptualization, Formal analysis, Funding acquisition, Investigation, Methodology, Project administration, Resources, Supervision, Writing – review \& editing;

\section*{Acknowledgment}
\noindent
Daniele Molino is a Ph.D. student enrolled in the National Ph.D. in Artificial Intelligence, XL cycle, course on Health and life sciences, organized by Università Campus Bio-Medico di Roma.
\\
This work was partially funded by: 
i) Università Campus Bio-Medico di Roma under the program ``University Strategic Projects'' within the project ``AI-powered Digital Twin for next-generation lung cancEr cAre (IDEA)''; 
ii) PNRR MUR project PE0000013-FAIR.
iii)  Cancerforskningsfonden Norrland project MP23-1122;
iv) Kempe Foundation project JCSMK24-0094; 
v) the Italian Ministry of Foreign Affairs and International Cooperation, grant number PGR01156
\\
Resources are provided by the National Academic Infrastructure for Supercomputing in Sweden (NAISS) and the Swedish National Infrastructure for Computing (SNIC) at Alvis @ C3SE, partially funded by the Swedish Research Council through grant agreements no. 2022-06725 and no. 2018-05973.

\bibliographystyle{unsrt}
\bibliography{biblio.bib}

\begin{thebibliography}{10}

\bibitem{alowais2023revolutionizing}
Shuroug~A Alowais, Sahar~S Alghamdi, Nada Alsuhebany, Tariq Alqahtani, Abdulrahman~I Alshaya, Sumaya~N Almohareb, Atheer Aldairem, Mohammed Alrashed, Khalid Bin~Saleh, Hisham~A Badreldin, et~al.
\newblock {Revolutionizing healthcare: the role of artificial intelligence in clinical practice}.
\newblock {\em BMC medical education}, 23(1):689, 2023.

\bibitem{guarrasi2025systematic}
Valerio Guarrasi, Fatih Aksu, Camillo~Maria Caruso, Francesco Di~Feola, Aurora Rofena, Filippo Ruffini, and Paolo Soda.
\newblock A systematic review of intermediate fusion in multimodal deep learning for biomedical applications.
\newblock {\em Image and Vision Computing}, page 105509, 2025.

\bibitem{alzubaidi2023survey}
Laith Alzubaidi, Jinshuai Bai, Aiman Al-Sabaawi, Jose Santamar{\'\i}a, Ahmed~Shihab Albahri, Bashar Sami~Nayyef Al-dabbagh, Mohammed~A Fadhel, Mohamed Manoufali, Jinglan Zhang, Ali~H Al-Timemy, et~al.
\newblock {A survey on deep learning tools dealing with data scarcity: definitions, challenges, solutions, tips, and applications}.
\newblock {\em Journal of Big Data}, 10(1):46, 2023.

\bibitem{gdpr2016general}
General Data Protection~Regulation GDPR.
\newblock {General data protection regulation}.
\newblock {\em Regulation (EU) 2016/679 of the European Parliament and of the Council of 27 April 2016 on the protection of natural persons with regard to the processing of personal data and on the free movement of such data, and repealing Directive 95/46/EC}, 2016.

\bibitem{act1996health}
Accountability Act.
\newblock {Health insurance portability and accountability act of 1996}.
\newblock {\em Public law}, 104:191, 1996.

\bibitem{goodfellow2020generative}
Ian Goodfellow, Jean Pouget-Abadie, Mehdi Mirza, Bing Xu, David Warde-Farley, Sherjil Ozair, Aaron Courville, and Yoshua Bengio.
\newblock {Generative adversarial networks}.
\newblock {\em Communications of the ACM}, 63(11):139--144, 2020.

\bibitem{saad2024survey}
Muhammad~Muneeb Saad, Ruairi O’Reilly, and Mubashir~Husain Rehmani.
\newblock A survey on training challenges in generative adversarial networks for biomedical image analysis.
\newblock {\em Artificial Intelligence Review}, 57(2):19, 2024.

\bibitem{ho2020denoising}
Jonathan Ho, Ajay Jain, and Pieter Abbeel.
\newblock {Denoising diffusion probabilistic models}.
\newblock In {\em Proceedings of the 34th International Conference on Neural Information Processing Systems}, NIPS '20. Curran Associates Inc., 2020.

\bibitem{rombach2022high}
Robin Rombach, Andreas Blattmann, Dominik Lorenz, Patrick Esser, and Bj\"orn Ommer.
\newblock {High-Resolution Image Synthesis With Latent Diffusion Models}.
\newblock In {\em Proceedings of the IEEE/CVF Conference on Computer Vision and Pattern Recognition (CVPR)}, pages 10684--10695, June 2022.

\bibitem{ramesh2022hierarchical}
Aditya Ramesh, Prafulla Dhariwal, Alex Nichol, Casey Chu, and Mark Chen.
\newblock {Hierarchical text-conditional image generation with clip latents}.
\newblock {\em arXiv preprint arXiv:2204.06125}, 1(2):3, 2022.

\bibitem{saharia2022photorealistic}
Chitwan Saharia, William Chan, Saurabh Saxena, Lala Li, Jay Whang, Emily~L Denton, Kamyar Ghasemipour, Raphael Gontijo~Lopes, Burcu Karagol~Ayan, Tim Salimans, et~al.
\newblock {Photorealistic text-to-image diffusion models with deep language understanding}.
\newblock {\em Advances in neural information processing systems}, 35:36479--36494, 2022.

\bibitem{bommasani2021opportunities}
Rishi Bommasani, Drew~A Hudson, Ehsan Adeli, Russ Altman, Simran Arora, Sydney von Arx, Michael~S Bernstein, Jeannette Bohg, Antoine Bosselut, Emma Brunskill, et~al.
\newblock {On the opportunities and risks of foundation models}.
\newblock {\em arXiv preprint arXiv:2108.07258}, 2021.

\bibitem{tang2023anytoany}
Zineng Tang, Ziyi Yang, Chenguang Zhu, Michael Zeng, and Mohit Bansal.
\newblock {Any-to-Any Generation via Composable Diffusion}.
\newblock In {\em Thirty-seventh Conference on Neural Information Processing Systems}, 2023.

\bibitem{li2023error}
Yangming Li and Mihaela van~der Schaar.
\newblock On error propagation of diffusion models.
\newblock {\em arXiv preprint arXiv:2308.05021}, 2023.

\bibitem{chen2020generating}
Zhihong Chen, Yan Song, Tsung-Hui Chang, and Xiang Wan.
\newblock Generating radiology reports via memory-driven transformer.
\newblock {\em arXiv preprint arXiv:2010.16056}, 2020.

\bibitem{nicolson2024longitudinal}
Aaron Nicolson, Jason Dowling, Douglas Anderson, and Bevan Koopman.
\newblock Longitudinal data and a semantic similarity reward for chest x-ray report generation.
\newblock {\em Informatics in Medicine Unlocked}, 50:101585, 2024.

\bibitem{miura2020improving}
Yasuhide Miura, Yuhao Zhang, Emily~Bao Tsai, Curtis~P Langlotz, and Dan Jurafsky.
\newblock Improving factual completeness and consistency of image-to-text radiology report generation.
\newblock {\em arXiv preprint arXiv:2010.10042}, 2020.

\bibitem{moris2022unsupervised}
Daniel~I Mor{\'\i}s, Joaquim de~Moura, Jorge Novo, and Marcos Ortega.
\newblock {Unsupervised contrastive unpaired image generation approach for improving tuberculosis screening using chest X-ray images}.
\newblock {\em Pattern Recognition Letters}, 164:60--66, 2022.

\bibitem{srivastav2021improved}
Devansh Srivastav, Akansha Bajpai, and Prakash Srivastava.
\newblock {Improved classification for pneumonia detection using transfer learning with GAN based synthetic image augmentation}.
\newblock In {\em 2021 11th international conference on cloud computing, data science \& engineering (confluence)}, pages 433--437. IEEE, 2021.

\bibitem{karbhari2021generation}
Yash Karbhari, Arpan Basu, Zong~Woo Geem, Gi-Tae Han, and Ram Sarkar.
\newblock {Generation of synthetic chest X-ray images and detection of COVID-19: A deep learning based approach}.
\newblock {\em Diagnostics}, 11(5):895, 2021.

\bibitem{shams2020generative}
MY~Shams, OM~Elzeki, Mohamed Abd~Elfattah, T~Medhat, and Aboul~Ella Hassanien.
\newblock {Why are generative adversarial networks vital for deep neural networks? A case study on COVID-19 chest X-ray images}.
\newblock In {\em Big data analytics and artificial intelligence against COVID-19: innovation vision and approach}, pages 147--162. Springer, 2020.

\bibitem{chambon2022roentgen}
Christian Bluethgen, Pierre Chambon, Jean-Benoit Delbrouck, Rogier van~der Sluijs, Ma{\l}gorzata Po{\l}acin, Juan~Manuel Zambrano~Chaves, Tanishq~Mathew Abraham, Shivanshu Purohit, Curtis~P Langlotz, and Akshay~S Chaudhari.
\newblock A vision--language foundation model for the generation of realistic chest x-ray images.
\newblock {\em Nature Biomedical Engineering}, pages 1--13, 2024.

\bibitem{chambon2022adapting}
Pierre Chambon, Christian Bluethgen, Curtis~P Langlotz, and Akshay Chaudhari.
\newblock {Adapting pretrained vision-language foundational models to medical imaging domains}.
\newblock {\em arXiv preprint arXiv:2210.04133}, 2022.

\bibitem{packhauser2023generation}
Kai Packh{\"a}user, Lukas Folle, Florian Thamm, and Andreas Maier.
\newblock {Generation of anonymous chest radiographs using latent diffusion models for training thoracic abnormality classification systems}.
\newblock In {\em 2023 IEEE 20th International Symposium on Biomedical Imaging (ISBI)}, pages 1--5. IEEE, 2023.

\bibitem{lee2023unified}
Hyungyung Lee, Wonjae Kim, Jin-Hwa Kim, Tackeun Kim, Jihang Kim, Leonard Sunwoo, and Edward Choi.
\newblock {Unified chest x-ray and radiology report generation model with multi-view chest x-rays}.
\newblock {\em arXiv preprint arXiv:2302.12172}, 3(7):8, 2023.

\bibitem{vaswani2017attention}
A~Vaswani.
\newblock {Attention is all you need}.
\newblock {\em Advances in Neural Information Processing Systems}, 2017.

\bibitem{choromanski2020rethinking}
Krzysztof Choromanski, Valerii Likhosherstov, David Dohan, Xingyou Song, Andreea Gane, Tamas Sarlos, Peter Hawkins, Jared Davis, Afroz Mohiuddin, Lukasz Kaiser, et~al.
\newblock {Rethinking attention with performers}.
\newblock {\em arXiv preprint arXiv:2009.14794}, 2020.

\bibitem{esser2021taming}
Patrick Esser, Robin Rombach, and Bjorn Ommer.
\newblock {Taming transformers for high-resolution image synthesis}.
\newblock In {\em Proceedings of the IEEE/CVF conference on computer vision and pattern recognition}, pages 12873--12883, 2021.

\bibitem{lee2023llm}
Suhyeon Lee, Won~Jun Kim, Jinho Chang, and Jong~Chul Ye.
\newblock {LLM-CXR: Instruction-Finetuned LLM for CXR Image Understanding and Generation}.
\newblock {\em arXiv preprint arXiv:2305.11490}, 2023.

\bibitem{oord2018representation}
Aaron van~den Oord, Yazhe Li, and Oriol Vinyals.
\newblock {Representation learning with contrastive predictive coding}.
\newblock {\em arXiv preprint arXiv:1807.03748}, 2018.

\bibitem{caruso2024maria}
Camillo~Maria Caruso, Paolo Soda, and Valerio Guarrasi.
\newblock Maria: a multimodal transformer model for incomplete healthcare data.
\newblock {\em arXiv preprint arXiv:2412.14810}, 2024.

\bibitem{johnson2019mimic}
Alistair~EW Johnson, Tom~J Pollard, Seth~J Berkowitz, Nathaniel~R Greenbaum, Matthew~P Lungren, Chih-ying Deng, Roger~G Mark, and Steven Horng.
\newblock {MIMIC-CXR, a de-identified publicly available database of chest radiographs with free-text reports}.
\newblock {\em Scientific data}, 6(1):317, 2019.

\bibitem{santosh2018angular}
KC~Santosh and Laurent Wendling.
\newblock {Angular relational signature-based chest radiograph image view classification}.
\newblock {\em Medical \& biological engineering \& computing}, 56:1447--1458, 2018.

\bibitem{raoof2012interpretation}
Suhail Raoof, David Feigin, Arthur Sung, Sabiha Raoof, Lavanya Irugulpati, and Edward~C Rosenow~III.
\newblock {Interpretation of plain chest roentgenogram}.
\newblock {\em Chest}, 141(2):545--558, 2012.

\bibitem{hashir2020quantifying}
Mohammad Hashir, Hadrien Bertrand, and Joseph~Paul Cohen.
\newblock {Quantifying the value of lateral views in deep learning for chest x-rays}.
\newblock In {\em Medical Imaging with Deep Learning}, pages 288--303. PMLR, 2020.

\bibitem{bidgood1997understanding}
W~Dean Bidgood~Jr, Steven~C Horii, Fred~W Prior, and Donald~E Van~Syckle.
\newblock {Understanding and using DICOM, the data interchange standard for biomedical imaging}.
\newblock {\em Journal of the American Medical Informatics Association}, 4(3):199--212, 1997.

\bibitem{radford2021learning}
Alec Radford, Jong~Wook Kim, Chris Hallacy, Aditya Ramesh, Gabriel Goh, Sandhini Agarwal, Girish Sastry, Amanda Askell, Pamela Mishkin, Jack Clark, et~al.
\newblock {Learning transferable visual models from natural language supervision}.
\newblock In {\em International conference on machine learning}, pages 8748--8763. PMLR, 2021.

\bibitem{li2020optimus}
Chunyuan Li, Xiang Gao, Yuan Li, Baolin Peng, Xiujun Li, Yizhe Zhang, and Jianfeng Gao.
\newblock {Optimus: Organizing sentences via pre-trained modeling of a latent space}.
\newblock {\em arXiv preprint arXiv:2004.04092}, 2020.

\bibitem{devlin2018bert}
Jacob Devlin Ming-Wei~Chang Kenton and Lee~Kristina Toutanova.
\newblock Bert: Pre-training of deep bidirectional transformers for language understanding.
\newblock In {\em Proceedings of naacL-HLT}, volume~1, page~2. Minneapolis, Minnesota, 2019.

\bibitem{radford2019language}
Alec Radford, Jeffrey Wu, Rewon Child, David Luan, Dario Amodei, Ilya Sutskever, et~al.
\newblock {Language models are unsupervised multitask learners}.
\newblock {\em OpenAI blog}, 1(8):9, 2019.

\bibitem{xu2023versatile}
Xingqian Xu, Zhangyang Wang, Gong Zhang, Kai Wang, and Humphrey Shi.
\newblock {Versatile diffusion: Text, images and variations all in one diffusion model}.
\newblock In {\em Proceedings of the IEEE/CVF International Conference on Computer Vision}, pages 7754--7765, 2023.

\bibitem{wu2018group}
Yuxin Wu and Kaiming He.
\newblock {Group normalization}.
\newblock In {\em Proceedings of the European conference on computer vision (ECCV)}, pages 3--19, 2018.

\bibitem{elfwing2018sigmoid}
Stefan Elfwing, Eiji Uchibe, and Kenji Doya.
\newblock {Sigmoid-weighted linear units for neural network function approximation in reinforcement learning}.
\newblock {\em Neural networks}, 107:3--11, 2018.

\bibitem{molino2025any}
Daniele Molino, Francesco di~Feola, Linlin Shen, Paolo Soda, and Valerio Guarrasi.
\newblock Any-to-any vision-language model for multimodal x-ray imaging and radiological report generation.
\newblock {\em arXiv preprint arXiv:2505.01091}, 2025.

\bibitem{boecking2022making}
Benedikt Boecking, Naoto Usuyama, Shruthi Bannur, Daniel~C Castro, Anton Schwaighofer, Stephanie Hyland, Maria Wetscherek, Tristan Naumann, Aditya Nori, Javier Alvarez-Valle, et~al.
\newblock Making the most of text semantics to improve biomedical vision--language processing.
\newblock In {\em European conference on computer vision}, pages 1--21. Springer, 2022.

\bibitem{heusel2017gans}
Martin Heusel, Hubert Ramsauer, Thomas Unterthiner, Bernhard Nessler, and Sepp Hochreiter.
\newblock {Gans trained by a two time-scale update rule converge to a local nash equilibrium}.
\newblock {\em Advances in neural information processing systems}, 30, 2017.

\bibitem{russakovsky2015imagenet}
Olga Russakovsky, Jia Deng, Hao Su, Jonathan Krause, Sanjeev Satheesh, Sean Ma, Zhiheng Huang, Andrej Karpathy, Aditya Khosla, Michael Bernstein, et~al.
\newblock {Imagenet large scale visual recognition challenge}.
\newblock {\em International journal of computer vision}, 115:211--252, 2015.

\bibitem{tronchin2021evaluating}
Lorenzo Tronchin, Rosa Sicilia, Ermanno Cordelli, Sara Ramella, and Paolo Soda.
\newblock E{valuating GANs in medical imaging}.
\newblock In {\em Deep Generative Models, and Data Augmentation, Labelling, and Imperfections: First Workshop, DGM4MICCAI 2021, and First Workshop, DALI 2021, Held in Conjunction with MICCAI 2021, Strasbourg, France, October 1, 2021, Proceedings 1}, pages 112--121. Springer, 2021.

\bibitem{cohen2022torchxrayvision}
Joseph~Paul Cohen, Joseph~D Viviano, Paul Bertin, Paul Morrison, Parsa Torabian, Matteo Guarrera, Matthew~P Lungren, Akshay Chaudhari, Rupert Brooks, Mohammad Hashir, et~al.
\newblock {TorchXRayVision: A library of chest X-ray datasets and models}.
\newblock In {\em International Conference on Medical Imaging with Deep Learning}, pages 231--249. PMLR, 2022.

\bibitem{papineni2002bleu}
Kishore Papineni, Salim Roukos, Todd Ward, and Wei-Jing Zhu.
\newblock {Bleu: a method for automatic evaluation of machine translation}.
\newblock In {\em Proceedings of the 40th annual meeting of the Association for Computational Linguistics}, pages 311--318, 2002.

\bibitem{irvin2019chexpert}
Jeremy Irvin, Pranav Rajpurkar, Michael Ko, Yifan Yu, Silviana Ciurea-Ilcus, Chris Chute, Henrik Marklund, Behzad Haghgoo, Robyn Ball, Katie Shpanskaya, et~al.
\newblock {Chexpert: A large chest radiograph dataset with uncertainty labels and expert comparison}.
\newblock In {\em Proceedings of the AAAI conference on artificial intelligence}, volume~33, pages 590--597, 2019.

\bibitem{delbrouck2022improving}
Jean-Benoit Delbrouck, Pierre Chambon, Christian Bluethgen, Emily Tsai, Omar Almusa, and Curtis~P Langlotz.
\newblock Improving the factual correctness of radiology report generation with semantic rewards.
\newblock {\em arXiv preprint arXiv:2210.12186}, 2022.

\bibitem{zhao2024ratescore}
Weike Zhao, Chaoyi Wu, Xiaoman Zhang, Ya~Zhang, Yanfeng Wang, and Weidi Xie.
\newblock Ratescore: A metric for radiology report generation.
\newblock {\em arXiv preprint arXiv:2406.16845}, 2024.

\bibitem{hamming1950error}
Richard~W Hamming.
\newblock {Error detecting and error correcting codes}.
\newblock {\em The Bell system technical journal}, 29(2):147--160, 1950.

\end{thebibliography}

\clearpage
\renewcommand{\thesection}{A\arabic{section}}
\renewcommand{\thetable}{A\arabic{table}}
\renewcommand{\thefigure}{A\arabic{figure}}
\section*{\centering Appendix}
\setcounter{section}{0}
\setcounter{table}{0}
\setcounter{figure}{0}

\section{FID Score computed with the XRV-mimic backbone}
\label{appendix:a}
\begin{table}[h]
\centering
\begin{adjustbox}{width=\textwidth}
\begin{tabular}{lcccccc}
\toprule
\textbf{Model} & \textbf{T$\rightarrow$F} & \textbf{L$\rightarrow$F} & \textbf{L$+$T$\rightarrow$F} & \textbf{T$\rightarrow$L} & \textbf{F$\rightarrow$L} & \textbf{F$+$T$\rightarrow$L} \\
& \textbf{XRV-mimic↓} & \textbf{XRV-mimic↓} & \textbf{XRV-mimic↓} & \textbf{XRV-mimic↓} & \textbf{XRV-mimic↓} & \textbf{XRV-mimic↓} \\
\midrule
\multirow{6}{*}{\begin{tabular}[c]{@{}l@{}} RoentGen \\ UniXGen \\ LLM-CXR \\ CoDi \\ $\text{CoDi}_{\mathit{{XR}}}$ \\ XGeM \end{tabular}}       
& 9.38 & - & -  & - & - & -  \\
& 17.08 & - & 16.36 & 22.12 & - & 23.07 \\
& 9.66 & - & -  & - & - & -  \\
& 141.79 & 140.03 & 141.53 & 139.92 & 142.34 & 139.13 \\
& \textbf{1.63} & 2.04 & 2.26 & \textbf{2.56} & 2.10 & 1.80 \\
& 1.68 & \textbf{0.45} & \textbf{0.44} & 2.78 & \textbf{0.48} & \textbf{0.44} \\
\bottomrule
\end{tabular}
\end{adjustbox}    
\captionsetup{justification=raggedright, singlelinecheck=false}
\caption{FID score for X-ray generation with XRV-mimic backbone, with lower values indicating greater similarity. The '-' symbol indicates that the respective models are not capable of performing the specified generation task. Results marked in bold denote the best performance.}
\label{tab:FID-xrvmimic}
\end{table}

\section{Intra-study BLEU score evaluation}
\label{appendix:b}
\begin{table}[h]
\centering
\begin{adjustbox}{width=0.6\textwidth}
\begin{tabular}{lcccc}
\toprule
\textbf{Model} & \textbf{BLEU-1} & \textbf{BLEU-2} & \textbf{BLEU-3} & \textbf{BLEU-4}\\
\multirow{3}{*}{\begin{tabular}[c]{@{}l@{}} UniXGen \\ $\text{CoDi}_{\mathit{{XR}}}$ \\ XGeM \end{tabular}}       
& .45 & .35 & .27 & .23 \\
& .59 & .52 & .48 & .46 \\
& \textbf{.60} & \textbf{.53} & \textbf{.49} & \textbf{.47} \\
\bottomrule
\end{tabular}
\end{adjustbox}
\caption{BLEU scores for intra-study evaluation of clinical report generation.
This setting assesses the model’s ability to generate consistent and semantically aligned reports for the same patient across multiple scans.
Higher scores indicate stronger intra-study coherence.
Bold values denote the highest performance for each column.}
\label{tab:bleu_study}
\end{table}

\clearpage

\section{Visual Turing Test}
\label{appendix:d}
\begin{table}[!h]
\centering
\resizebox{\textwidth}{!}{
\begin{tabular}{lcc}
\toprule
\textbf{\rr Task} & \textbf{\rr Data Type} & \textbf{\rr Score}\\
\midrule
\rr General X-ray Realism & \rr Real X-rays & \rr 4.1 ± 0.9 \\
                      & \rr Synthetic X-rays & \rr 3.7 ± 0.8 \\
\midrule
\rr General Report Realism & \rr Real Reports & \rr 3.5 ± 0.9 \\
                    & \rr Synthetic Reports & \rr 4.0 ± 0.9 \\
\midrule
\rr Report Coherence with X-ray Pair & \rr Real Reports & \rr 3.4 ± 1.3 \\
                                 & \rr Synthetic Reports & \rr 3.3 ± 1.1 \\
\midrule
\rr Coherence Between Report and X-ray & \rr Real X-rays & \rr 3.7 ± 0.9 \\
                                  & \rr Synthetic X-rays & \rr 3.9 ± 0.8 \\
\midrule
\rr Coherence Between X-ray Pairs & \rr Real X-rays & \rr 3.8 ± 0.7 \\
                            & \rr Synthetic X-rays & \rr 3.6 ± 0.8 \\
\bottomrule
\end{tabular}
}
\captionsetup{justification=raggedright, singlelinecheck=false}
\caption{Evaluation results for different tasks in the Visual Turing Test, comparing real and synthetic data generated by XGeM. The scores range from 0 to 5, with higher values indicating better performance.}
\label{tab:VTT_results}
\end{table}

\clearpage

\section{Anonymization, Imbalance Learning and Data Scarcity assessment with Lateral X-rays}
\label{appendix:c}
\begin{table}[h]
\centering
\begin{adjustbox}{width=0.8\textwidth}
\begin{tabular}{lcccccc}
\toprule
\textbf{Training Data} & \multicolumn{3}{c}{\textbf{AUROC}} & \multicolumn{3}{c}{\textbf{F1-Score}} \\ 
\cmidrule(lr){2-4} \cmidrule(lr){5-7}
& \textbf{Micro} & \textbf{Macro} & \textbf{Weighted} & \textbf{Micro} & \textbf{Macro} & \textbf{Weighted} \\
\midrule
\multirow{2}{*}{\begin{tabular}[c]{@{}l@{}} Real \\ Synthetic \end{tabular}}       
& .77 & .73 & .73 & .34 & .23 & .24 \\
& .75 & .70 & .69 & .37 & .26 & .27 \\
\bottomrule
\end{tabular}
\end{adjustbox}
\caption{Classification performance of two DenseNet-121 models trained respectively on real lateral CXR and on a combination of real and synthetic lateral CXR, evaluated on a held-out test set of real X-rays. Metrics include AUROC and F1-score. This experiment simulates an anonymization scenario, assessing whether synthetic data generated by XGeM can effectively substitute real data for model training without compromising performance.}\label{tab:Anonymization-Lat}
\end{table}
\begin{table}[h]
\centering
\begin{adjustbox}{width=0.8\textwidth}
\begin{tabular}{lcccccccccc}
\toprule
\textbf{Training Data} & \multicolumn{8}{c}{\textbf{F1-Score}} \\ 
\cmidrule(lr){2-9}
& \textbf{Atl.} & \textbf{Cmgl.} & \textbf{Cnsl.} & \textbf{Edm.} & \textbf{Eff.} & \textbf{Micro} & \textbf{Macro} & \textbf{Weighted} \\
\midrule
Real & .32 & .41 & .13 & .70 & .63 & .46 & .40 & .41 \\
Synthetic & .34 & .43 & .15 & .65 & .63 & .48 & .42 & .43 \\
\bottomrule
\end{tabular}
\end{adjustbox}
\caption{F1-score performance of DenseNet-121 classifiers trained on an imbalanced dataset of lateral CXR versus a balanced dataset obtained through XGeM–generated synthetic data. Each column reports the per-class F1-score computed on the real held-out test set. This experiment evaluates whether synthetic data can mitigate class imbalance and improve sensitivity to underrepresented but clinically significant conditions.}
\label{tab:Data-imbalance-lat}
\end{table}

\begin{table}[t]
\centering
\resizebox{\textwidth}{!}{
\begin{tabular}{ccc|ccc|ccc}
\toprule
\textbf{Real} & \textbf{Synthetic} & \textbf{Total} & \textbf{AUROC (Micro)} & \textbf{AUROC (Macro)} & \textbf{AUROC (Weighted)} & \textbf{F1 (Micro)} & \textbf{F1 (Macro)} & \textbf{F1 (Weighted)} \\
\midrule
15k  & 0     & 15k   & 0.68 & 0.65 & 0.66 & 0.22 & 0.15 & 0.16 \\
15k  & 5k    & 20k   & 0.70 & 0.67 & 0.68 & 0.25 & 0.17 & 0.18 \\
15k   & 15k   & 30k   & 0.72 & 0.70 & 0.70 & 0.28 & 0.20 & 0.21 \\
15k & 30k & 45k   & 0.74 & 0.72 & 0.72 & 0.31 & 0.22 & 0.23 \\
15k    & 60k   & 75k   & 0.74 & 0.73 & 0.73 & 0.34 & 0.23 & 0.24 \\
\bottomrule
\end{tabular}
}
\caption{Classification performance of DenseNet-121 models trained on subsets of 15k real lateral CXRs augmented with increasing amounts of synthetic data from XGeM. Metrics are reported on the same real test set.}
\label{tab:Data-scarcity-lat}
\end{table}

\end{document}